\documentclass{article}

\usepackage[utf8]{inputenc}
\usepackage{url}
\usepackage[T1]{fontenc}
\usepackage{amsmath}
\usepackage{amsthm}
\usepackage{amsfonts}
\usepackage{amssymb}

\usepackage{xcolor}
\usepackage{enumitem}

\def\eg{\emph{e.g.}}
\def\ie{\emph{i.e.}}
\def\Hcal{{\mathcal H}}

\def\Xcal{{\mathcal X}}
\def\Real{{\mathbb R}}

\def\R{{\mathbb R}}

\def\Hc{{\mathcal H}}

\DeclareMathOperator{\E}{\mathbb{E}}

\newtheorem{theorem}{Theorem}
\newtheorem{proposition}[theorem]{Proposition}

\usepackage{microtype}
\usepackage{graphicx}
\usepackage{subfigure}
\usepackage{booktabs}
\usepackage{hyperref}

\usepackage[accepted]{icml2019}

\icmltitlerunning{A Kernel Perspective for Regularizing Deep Neural Networks}

\begin{document}

\twocolumn[
\icmltitle{A Kernel Perspective for Regularizing Deep Neural Networks}

\icmlsetsymbol{equal}{*}

\begin{icmlauthorlist}
\icmlauthor{Alberto Bietti}{equal,thoth}
\icmlauthor{Grégoire Mialon}{equal,thoth,sierra}
\icmlauthor{Dexiong Chen}{thoth}
\icmlauthor{Julien Mairal}{thoth}

\end{icmlauthorlist}

\icmlaffiliation{thoth}{Univ. Grenoble Alpes, Inria, CNRS, Grenoble INP, LJK, 38000 Grenoble, France}
\icmlaffiliation{sierra}{Département d'informatique de l'ENS, ENS, CNRS, Inria, PSL, 75005 Paris, France}

\icmlcorrespondingauthor{}{firstname.lastname@inria.fr}

\icmlkeywords{machine learning, kernel methods, regularization, rkhs, deep learning}

\vskip 0.3in
]

\printAffiliationsAndNotice{\icmlEqualContribution}

\begin{abstract}
We propose a new point of view for regularizing deep neural networks by using 
the norm of a reproducing kernel Hilbert space (RKHS). Even though this norm cannot be computed,
it admits upper and lower approximations leading to various practical strategies. 
Specifically, this perspective (i) provides a common umbrella for many existing regularization principles, including spectral norm and gradient penalties,
or adversarial training, (ii) leads to new effective regularization penalties, and
(iii) suggests hybrid strategies combining lower and upper bounds to get better approximations of the RKHS norm.
We experimentally show this approach
to be effective when learning on small datasets, or to obtain
adversarially robust models.

\end{abstract}

\vspace{-0.3cm}
\section{Introduction}
\label{sec:introduction}

Learning predictive models for complex tasks often requires large amounts of
annotated data. For instance, convolutional neural networks are
huge-dimensional and typically involve more parameters than training samples,
which raises several challenges: achieving good generalization with small
datasets is indeed difficult, which limits the deployment of such deep models to
many tasks where labeled data is scarce, \eg, in biology~\citep{ching2018opportunities}.
Besides, imperceptible adversarial perturbations can
significantly degrade the prediction quality~\citep{szegedy2013intriguing,biggio2018wild}.
These issues raise the question of regularization as an essential
tool to control the complexity of deep models, as well as their stability
to small variations of their inputs.

In this paper, we present a new perspective on regularization of deep networks,
by viewing convolutional neural networks (CNNs) as elements of a
RKHS following the work
of~\citet{bietti2018group} on deep convolutional kernels. 
For such kernels, the RKHS contains indeed deep convolutional networks similar
to generic ones---up to smooth approximations of rectified linear units.
Such a point of view provides a
natural regularization function, the RKHS norm, which allows us to control the
variations of the predictive model and
to limit its complexity for better generalization.
Besides, the norm also acts as a Lipschitz constant, which provides a
direct control on the stability to adversarial perturbations.

In contrast to traditional kernel methods, the RKHS norm cannot be explicitly
computed in our setup. Yet, this norm admits numerous approximations---lower
bounds and upper bounds---which lead to many strategies for regularization
based on penalties, constraints, or combinations thereof.  Depending on the
chosen approximation, we recover then many existing principles such as spectral norm
regularization~\citep{cisse2017parseval,yoshida2017spectral,miyato2018spectral,sedghi2018singular},
gradient penalties and double backpropagation~\citep{drucker1991double,simon2018adversarial,gulrajani2017improved,roth2017stabilizing,roth2018adversarially,arbel2018gradient},
adversarial training~\citep{madry2018towards},
and we also draw links with tangent
propagation~\citep{simard1998transformation}.
For all these principles, we provide a unified viewpoint and
theoretical insights, and we also introduce new variants, which we show
are effective in practice when learning with few labeled data, or in 
the presence of adversarial perturbations.

Moreover, regularization and robustness are tightly linked in our kernel framework.
Specifically, some lower bounds on the RKHS norm lead to robust optimization objectives with worst-case $\ell_2$ perturbations;
further, we can extend margin-based generalization bounds in the spirit
of~\citet{bartlett2017spectrally,boucheron2005theory} to the setting of \emph{adversarially robust}
generalization~\citep[see][]{schmidt2018adversarially}, where an adversary can perturb test data.
We also discuss connections between recent regularization strategies for training generative adversarial networks
and approaches to generative modeling based on kernel two-sample tests (MMD)~\citep{dziugaite2015training,li2017mmd,binkowski2018demystifying}.

\vspace{-0.2cm}
\paragraph{Summary of the contributions.} ~ \\
~$\bullet$~ We introduce an RKHS perspective for regularizing deep neural networks models which provides
	a unified view on various practical regularization principles,
	together with theoretical insight and guarantees;\\
	~$\bullet$~ By considering lower bounds to the RKHS norm, we obtain new penalties based on adversarial perturbations,
	adversarial deformations, or gradient norms of prediction functions, which we show to be effective in practice; \\
	~$\bullet$~ Our RKHS point of view suggests combined strategies based on both upper and
	lower bounds, which we show often perform empirically best in the context of generalization from small image and biological datasets,
	by providing a tighter control of the RKHS norm.

\vspace{-0.2cm}
\paragraph{Related work.}
The construction of hierarchical kernels and the study of neural networks in the corresponding RKHS was studied
by~\citet{mairal2016end,zhang2016l1,zhang2016convexified,bietti2018group}.
Some of the regularization strategies we obtain from our kernel perspective are variants of previous
approaches to adversarial robustness~\citep{cisse2017parseval,madry2018towards,simon2018adversarial,roth2018adversarially},
to improving generalization~\citep{drucker1991double,miyato2018virtual,sedghi2018singular,simard1998transformation,yoshida2017spectral},
and stable training of generative adversarial networks~\citep{roth2017stabilizing,gulrajani2017improved,arbel2018gradient,miyato2018spectral}.
The link between robust optimization and regularization was studied by~\citet{xu2009robust,xu2009robustness},
focusing mainly on linear models with quadratic or hinge losses.
The notion of adversarial generalization was considered by~\citet{schmidt2018adversarially},
who provide lower bounds on a particular data distribution.
\citet{sinha2018certifying} provide generalization guarantees in the different setting of distributional robustness;
compared to our bound, they consider expected loss instead of classification error, and their bounds do not highlight the dependence on
the model complexity.

\section{Regularization of Deep Neural Networks}
\label{sec:kernel_reg}

In this section, we recall the kernel perspective on deep networks introduced
by~\citet{bietti2018group}, and present upper and lower bounds on the RKHS norm
of a given model, leading to various regularization strategies.  For
simplicity, we first consider real-valued networks and binary classification,
before discussing multi-class extensions.

\subsection{Relation between deep networks and RKHSs}
\label{sub:rkhs_construction}
Kernel methods consist of mapping data living in a set~$\Xcal$ to a
RKHS~$\Hcal$ associated to a positive definite kernel~$K$ through a mapping
function $\Phi: \Xcal \to \Hcal$, and then learning simple machine learning
models in~$\Hcal$. Specifically, when considering a real-valued regression or
binary classification problem, classical kernel methods find a prediction
function $f : \mathcal{X} \to \R$ living in the RKHS which can be written in linear
form, i.e., such that $f(x) = \langle f, \Phi(x) \rangle_\Hc$ for all~$x$
in~$\Xcal$.  While explicit mapping to a possibly infinite-dimensional space is
of course only an abstract mathematical operation, learning~$f$ can be done
implicitly by computing kernel evaluations and typically by using convex
programming~\citep{scholkopf2001learning}.

Moreover, the RKHS norm~$\|f\|_\Hc$ acts as a natural regularization function,
which controls the variations of model predictions
according to the geometry induced by~$\Phi$:
\begin{equation}
\label{eq:cs}
|f(x) - f(x')| \leq \|f\|_\Hc \cdot \|\Phi(x) - \Phi(x') \|_\Hc.
\end{equation}
Unfortunately, our setup does not allow us to use the RKHS norm in a traditional way since evaluating the kernel is intractable. Instead, we
propose a different approach that considers explicit parameterized
representations of functions contained in the RKHS, given by generic CNNs,
and leverage properties of the RKHS and the
kernel mapping in order to regularize when learning the network parameters.

Consider indeed a real-valued deep convolutional network $f : \mathcal{X} \to
\R$, where $\mathcal X$ is simply $\R^d$, with rectified linear unit (ReLU) activations and no bias units.
By constructing an appropriate multi-layer hierarchical kernel, \citet{bietti2018group} show
that the corresponding RKHS~$\Hc$ contains a CNN with the same architecture and parameters as~$f$,
but with activations that are smooth approximations of ReLU.
Although the model predictions might not be strictly equal, we will abuse notation and denote this
approximation with smooth ReLU by~$f$ as well,
with the hope that the regularization procedures derived from the RKHS model
will be effective in practice on the original CNN~$f$.

Besides, the mapping~$\Phi(\cdot)$ is shown to be non-expansive:
\begin{equation}
\label{eq:non_expansive}
\|\Phi(x) - \Phi(x') \|_\Hc \leq \|x - x'\|_2,
\end{equation}
so that controlling~$\|f\|_\Hc$ provides some robustness to additive $\ell_2$-perturbations, by~\eqref{eq:cs}.
Additionally, with appropriate pooling operations, \citet{bietti2018group} show that the kernel mapping is also
stable to deformations, meaning that the RKHS norm also controls robustness to translations
and other transformations including scaling and rotations,
which can be seen as deformations when they are small.

In contrast to standard kernel methods, where the RKHS norm is typically available in closed form, this norm is difficult to compute in our setup, and requires approximations.
The following sections present upper and lower bounds on~$\|f\|_{\Hcal}$,
with linear convolutional operations denoted by~$W_k$ for $k=1, \ldots, L$, where~$L$ is the number of layers.
Defining~$\theta := \{W_k : k = 1, \ldots, L\}$, we then leverage these bounds to approximately solve the following
penalized or constrained optimization problems on a training set~$(x_i, y_i), i = 1, \ldots, n$:
\begin{align}
\label{eq:penalty_or_constraint}
& \min_\theta \frac{1}{n} \sum_{i=1}^n \ell(y_i, f_\theta(x_i)) + \lambda \|f_\theta\|_\Hc^2 \quad \text{or } \\
& \min_{\theta: \|f_\theta\|_\Hc \leq C} \frac{1}{n} \sum_{i=1}^n \ell(y_i, f_\theta(x_i)).
\end{align}
We also note that while the construction of~\citet{bietti2018group} considers VGG-like networks~\citep{simonyan2014very},
the regularization algorithms we obtain in practice can be easily adapted to different architectures
such as residual networks~\citep{he2016deep}.

\subsection{Exploiting lower bounds of the RKHS norm}
\label{sub:lower_bounds}

In this section, we devise regularization algorithms by leveraging lower bounds on~$\|f\|_\Hc$,
obtained by relying on the following variational characterization of Hilbert norms:
\begin{equation*}
\|f\|_\Hc = \sup_{\|u\|_\Hc \leq 1} \langle f, u \rangle_\Hc.
\end{equation*}
At first sight, this definition is not useful since the set $U = \{u \in \Hc : \|u\|_\Hc \leq 1\}$ may be 
infinite-dimensional
 and the inner products $\langle f, u \rangle_\Hc$ cannot be
computed in general. Thus, we devise tractable lower bound approximations by considering smaller sets~$\bar{U} \subset U$.

\paragraph{Adversarial perturbation penalty.}
Thanks to the non-expansiveness of $\Phi$, we can consider the subset $\bar U \subset U$ defined as $\bar{U} = \{\Phi(x + \delta) - \Phi(x) : x \in \mathcal X, \|\delta\|_2 \leq 1 \}$,
leading to the bound
\begin{equation}
\label{eq:lower_bound}
\|f\|_\Hc  \geq  \|f\|_\delta^2 := \sup_{x \in \mathcal X, \|\delta\|_2 \leq 1} f(x + \delta) - f(x),
\end{equation}
which is reminiscent of adversarial perturbations. Adding a regularization parameter $\epsilon > 0$ in front of the norm
then corresponds to different sizes of perturbations:
\begin{equation}
\label{eq:kernel_adv}
\epsilon \|f\|_\Hc = \sup_{\|u\|_\Hc \leq \epsilon} \langle f, u \rangle_\Hc \geq \sup_{x \in \mathcal X, \|\delta\|_2 \leq \epsilon} f(x + \delta) - f(x).
\end{equation}
Using this lower bound or its square as a penalty in the objective~\eqref{eq:penalty_or_constraint}
when training a CNN provides a way to regularize.
Optimizing over adversarial perturbations has been useful to obtain robust models~\citep[\eg, the PGD method of~][]{madry2018towards};
yet our approach differs in two important ways: 

(i) it involves a penalty that is decoupled from the loss term such that 
in principle, our penalty could be used beyond the supervised empirical risk paradigm.
In contrast, PGD optimizes the robust formulation~\eqref{eq:robust} below, which 
fits training data while considering
perturbations on the loss.

(ii) our penalty involves a global maximization problem
on the input space~$\Xcal$, as opposed to only maximizing on perturbations near
training data. In practice, optimizing over~$\Xcal$ is however
difficult and instead, we replace~$\Xcal$ by random mini-batches of examples,
yielding further lower bounds on the RKHS norm. These examples may be labeled or not,
in contrast to PGD that perturb labeled examples only.
When using such a mini-batch,
a gradient of the penalty can be obtained by first finding maximizers~$\hat x, \hat \delta$
(where~$\hat x$ is an element of the mini-batch and $\hat{\delta}$ is a perturbation), and then computing gradients
of $f_\theta(\hat x + \hat \delta) - f_\theta(\hat x)$ with respect to~$\theta$ by using back-propagation.
In practice, we compute the perturbations~$\delta$ for each example~$x$ by using a few steps of
projected gradient ascent with constant step-lengths.

\paragraph{Robust optimization yields another lower bound.}
In some contexts, our penalized approach is related to solving the robust optimization problem
\begin{equation}
\label{eq:robust}
\min_\theta \frac{1}{n} \sum_{i=1}^n \sup_{\|\delta\|_2 \leq \epsilon} \ell(y_i, f_\theta(x_i + \delta)),
\end{equation}
which is commonly considered for training adversarially robust classifiers~\citep{wong2018provable,madry2018towards,raghunathan2018certified}.
In particular, \citet{xu2009robustness} show that the penalized and
robust objectives are equivalent in the case of the hinge loss with linear predictors,
when the data is non-separable.
They also show the equivalence for kernel methods when considering the (intractable) full perturbation set~$U$
around each point in the RKHS~$\Phi(x_i)$, that is, predictions $\langle f, \Phi(x_i) + u \rangle_\Hc$ with~$u$ in $U$.
Intuitively, when a training example $(x_i, y_i)$ is misclassified, we are in the ``linear'' part of the hinge loss, such~that
\begin{equation*}
\sup_{\|u\|_\Hc \leq \epsilon} \ell(y_i, \langle f, \Phi(x_i) + u \rangle_\Hc) = \ell(y_i, f(x_i)) + \epsilon \|f\|_{\Hcal}.
\end{equation*}
For other losses such as the logistic loss, a regularization effect is still present even
for correctly classified examples,
though it may be smaller since the loss has a reduced slope for such points.
This leads to an \emph{adaptive} regularization mechanism that may automatically reduce
the amount of regularization when the data is easily separable.
However, the robust optimization approach might only encourage local stability around training examples, while the global quantity~$\|f\|_\Hc$
may become large in order to better fit the data.
We note that a perfect fit of the data with large complexity does not prevent generalization~\citep[see, \eg,][]{belkin2018overfitting,belkin2018understand};
yet, such mechanisms are still poorly understood.
Nevertheless, it is easy to show that the robust objective~\eqref{eq:robust}
lower bounds the penalized objective with penalty~$\epsilon \|f\|_{\Hcal}$.

\paragraph{Gradient penalties.}
Taking $\bar{U} \!= \! \{\frac{\Phi(x) - \Phi(y)}{\|x - y\|_2} : x, y \!\in\! \mathcal X\}$, which is a subset of~$U$
by Eq.~\eqref{eq:non_expansive}---it turns out that this is the same set as for adversarial perturbation penalties,
since~$\Phi$ is homogeneous~\citep{bietti2018group} and $\mathcal X =
\Real^d$---we obtain a lower bound based on the Lipschitz constant of~$f$:
\begin{equation}
\|f\|_\Hc \geq \sup_{x, y \in \mathcal{X}} \frac{f(x) - f(y)}{\|x - y\|_2} \geq \|\nabla f\| := \sup_{x \in \mathcal X} \|\nabla f(x)\|_2, \label{eq:gradientpenalty}
\end{equation}
where the second inequality becomes an equality when~$\Xcal$ is convex,
and the supremum is taken over points where~$f$ is differentiable.
Although we are unaware of previous work using this exact lower bound for a generic regularization penalty,
we note that variants replacing the supremum over~$x$ by an expectation over data have been recently used
to stabilize the training of generative adversarial networks~\citep{gulrajani2017improved,roth2017stabilizing},
and we provide insights in Section~\ref{sub:gan_reg} on the benefits of RKHS regularization in such a setting.
Related penalties have been considered in the context of robust optimization,
for regularization or robustness,
noting that a penalty based on the gradient of the loss function $x \mapsto \ell(y, f(x))$ can give a good approximation of~$\eqref{eq:robust}$
when~$\epsilon$ is small~\citep{drucker1991double,lyu2015unified,roth2018adversarially,simon2018adversarial}.

\paragraph{Penalties based on deformation stability.}
We may also obtain new penalties by considering more exotic sets
$\bar{U} = \{\Phi(\tilde x) - \Phi(x) : x~\in~{\mathcal X},~ \tilde x \text{ is a small} \text{ deformation of }x\}$,
where the amount of deformation is dictated by the stability bounds of~\citet{bietti2018group} in order to ensure that $\bar{U} \subset U$.
More precisely, such bounds depend on the maximum displacement and Jacobian norm of
the diffeomorphisms considered. These can be easily computed for various parameterized families
of transformations, such as translations, scaling or rotations, leading to simple ways to control
the regularization strength through the parameters of these transformations.
One can also consider infinitesimal deformations from such parameterized transformations,
which approximately yields the \emph{tangent propagation} regularization
strategy of~\citet{simard1998transformation}.
These approaches are detailed in Appendix~\ref{sec:deformation_penalties}.
If instead we consider the robust optimization formulation~\eqref{eq:robust}, we obtain a form
of \emph{data augmentation} where transformations are optimized instead of sampled, as done by~\citep{engstrom2017rotation}.

\paragraph{Extensions to multiple classes and beyond}
\label{sub:multiclass}

We now extend the regularization strategies based on lower bounds to multi-valued networks,
in order to deal with multiple classes.
For that purpose, we consider a multi-class penalty $\|f_1\|_\Hc^2 + \ldots + \|f_K\|_\Hc^2$
for an~$\R^K$-valued function $f = (f_1, f_2, \ldots, f_K)$, and we define 
\begin{align*}
\|f\|_\delta^2 := \sum_{k=1}^K \|f_k\|_\delta^2 \text{~~~and~~~} \|\nabla f\|^2 := \sum_{k=1}^K \|\nabla f_k\|^2,
\end{align*}
where $\|f_k\|_\delta$ is the adversarial penalty~(\ref{eq:lower_bound}), and $\|\nabla f_k\|$ is defined in~(\ref{eq:gradientpenalty}).
For deformation stability penalties, we proceed in a similar manner,
and for robust optimization formulations~\eqref{eq:robust}, the extension is straightforward,
given that multi-class losses such as cross-entropy can be
directly optimized in an adversarial training or gradient penalty setup.

Finally, we note that while the kernel approach we introduce considers
the Euclidian geometry in the input space, it is possible to consider heuristic alternatives for other
geometries, such as~$\ell_\infty$ perturbations, as discussed in Appendix~\ref{sec:non_euclidian_appx}.

\subsection{Exploiting upper bounds with spectral norms}
\label{sub:upper_bounds}

Instead of lower bounds, one may use instead
 the following upper bound from~\citet[Proposition 14]{bietti2018group}:
\begin{equation}
\label{eq:upper_bound}
\|f\|_\Hc \leq \omega(\|W_1\|, \ldots, \|W_L\|),
\end{equation}
where~$\omega$ is increasing in all of its arguments, and~$\|W_k\|$ is the spectral norm of the linear
operator~$W_k$.
Here, we simply consider the spectral norm on the filters, given by~$\|W\| := \sup_{\|x\|_2 \leq 1} \|W x\|_2$.
Other generalization bounds relying on similar quantities have been proposed for
controlling complexity~\citep{bartlett2017spectrally,neyshabur2017pac}, suggesting that using them for
regularization is relevant even beyond our kernel perspective,
as observed by~\citet{cisse2017parseval,sedghi2018singular,yoshida2017spectral}.
Extensions to multiple classes are simple to obtain by simply considering spectral norms up to
the last layer.

\paragraph{Penalizing the spectral norms.}
One way to control the upper bound~\eqref{eq:upper_bound} when learning a neural network~$f_\theta$
is to consider a regularization penalty based on spectral norms
\begin{equation} 
	\label{eq:optimization_problem_penalized} 
	\min_{\theta}\frac{1}{n} \sum_{i=1}^{n} \ell(y_i, f_{\theta}(x_i)) + \lambda \sum_{l=1}^{L} \|W_l\|^2,
\end{equation}
where~$\lambda$ is a regularization parameter.
To optimize this cost, one can obtain (sub)gradients of the penalty by computing singular vectors
associated to the largest singular value of each~$W_l$.
We consider the method of~\citet{yoshida2017spectral}, which computes such singular vectors approximately
using one or two iterations of the power method, as well as a more costly approach using the full SVD.

\paragraph{Constraining the spectral norms with a continuation approach.}
In the constrained setting, we want to optimize:
\begin{equation*}
\begin{aligned}
\min_{\theta}\frac{1}{n} \sum_{i=1}^{n} \ell(y_i, f_{\theta}(x_i)) \text{~~s.t.~~} \|W_l\| \leq \tau \text{ ; } l\in 1, \dots, L~,
\end{aligned}
\end{equation*}
where $\tau$ is a user-defined constraint.
This objective may be optimized by projecting each~$W_l$ in the
spectral norm ball of radius $\tau$ after each gradient step.
Such a projection is achieved by truncating the singular values to be smaller than~$\tau$ (see Appendix~\ref{sec:spectral_norms_appx}).
We found that the loss was hardly optimized with this approach,
and therefore introduce a continuation approach with an exponentially
decaying schedule for~$\tau$ reaching a constant $\tau_0$
after a few epochs, which we found to be important for good empirical performance.

\subsection{Combining upper and lower bounds.}
One advantage of lower bound penalties is that they are independent of
the model parameterization, making them flexible enough to use with more complex architectures.
In addition, the connection with robust optimization can provide a useful mechanism for adaptive regularization.
However, they do not provide a guaranteed control on the RKHS norm, unlike the upper bound strategies.
This is particularly true for robust optimization approaches, which may favor small training loss
and local stability over global stability through~$\|f\|_\Hc$.
Nevertheless, we observed that our new approaches based on separate penalties
sometimes do help in controlling upper bounds as well (see Section~\ref{sec:experiments}).

While these upper bound strategies are useful for limiting model complexity,
we found them empirically less effective for robustness (see Section~\ref{sub:exp_robust}).
However, we observed that combining with lower bound approaches can overcome this weakness,
perhaps due to a better control of local stability.
In particular, such combined approaches often provide the best generalization performance
in small data scenarios, as well as better guarantees on adversarially robust generalization
thanks to a tighter control of the RKHS norm.

\section{Theoretical Guarantees and Insights}
\label{sec:theory}

In this section, we study how the kernel perspective allows us to extend standard margin-based generalization bounds 
to an adversarial setting in order to
provide theoretical guarantees on adversarially robust generalization.
We then discuss how our kernel approach provides novel interpretations for training
generative adversarial networks.

\subsection{Guarantees on adversarial generalization}
\label{sub:guarantees}

While various methods have been introduced to empirically gain robustness to adversarial perturbations,
the ability to generalize with such perturbations, also known as \emph{adversarial generalization}~\citep{schmidt2018adversarially},
still lacks theoretical understanding.
Margin-based bounds have been useful
to explain the generalization behavior of learning algorithms that can fit the training data
well, such as kernel methods, boosting and neural networks~\citep{koltchinskii2002empirical,boucheron2005theory,bartlett2017spectrally}.
Here, we show how such arguments can be adapted to obtain guarantees on adversarial generalization,
\ie, on the expected classification error in the presence of an~$\ell_2$-bounded adversary,
based on the RKHS norm of a learned model.
For a binary classification task with labels in $\mathcal Y = \{-1,1\}$ and data distribution~$\mathcal D$,
we would like to bound the expected adversarial error of a classifier~$f$, given for some $\epsilon > 0$ by
\begin{equation}
\label{eq:adv_error}
\text{err}_\mathcal D(f, \epsilon) := P_{(x,y) \sim \mathcal D} (\exists \|\delta\|_2 \leq \epsilon: ~y f(x + \delta) < 0).
\end{equation}
Leveraging the fact that~$f$ is $\|f\|_\Hc$-Lipschitz,
we now show how to further bound this quantity using empirical margins,
following the usual approach to obtaining margin bounds for kernel methods~\citep[\eg,][]{boucheron2005theory}.
Consider a training dataset $(x_1, y_1), \ldots, (x_n, y_n) \in \mathcal X \times \mathcal Y$.
Defining $L_n^\gamma(f) := \frac{1}{n} \sum_{i=1}^n \textbf{1}\{y_i f(x_i) < \gamma\}$,
we have the following bound, proved in Appendix~\ref{sec:generalization_appx}:
\begin{proposition}[Adversarially robust margin bound]
\label{prop:robust_margin_bound}
With probability~$1 - \delta$ over a dataset $\{(x_i, y_i)\}_{i=1, \ldots, n}$, we have,
for all choices of $\gamma > 0$ and~$f \in \Hc$,
\begin{align}
\label{eq:robust_margin_bound}
& \text{err}_\mathcal D(f, \epsilon) \leq L_n^{\gamma + 2 \epsilon \|f\|_\Hc}(f) + \tilde{O}\left( \frac{\|f\|_\Hc \bar{B}}{\gamma \sqrt{n}}  \right),
\end{align}
where $\bar{B} = \sqrt{\frac{1}{n}\sum_{i=1}^n K(x_i, x_i)}$ and $\tilde{O}$ hides a term depending logarithmically on~$\|f\|_\Hc, \gamma$, and $\delta$.
\end{proposition}

When $\epsilon = 0$, we obtain the usual margin bound, while $\epsilon > 0$ yields
a bound on adversarial error~$\text{err}_\mathcal D(f, \epsilon)$,
for some neural network~$f$ learned from data.
Note that other complexity measures based on products of spectral norms may be used instead of~$\|f\|_\Hcal$,
as well as multi-class extensions, following~\citet{bartlett2017spectrally,neyshabur2017pac}.
In concurrent work, \citet{khim2018adversarial,yin2019rademacher} derive similar bounds
in the context of fully-connected networks.
In contrast to these works, which bound complexity of a modified function class,
our bound uses the complexity of the original class and leverages smoothness properties
of functions to derive the margin bound.

One can then study the effectiveness of a regularization algorithm by inspecting
cumulative distribution (CDF) plots of the
\emph{normalized margins} $\bar{\gamma}_i = y_i f(x_i) / \|f\|_\Hc$,
for different strengths of regularization (an example is given in Figure~\ref{fig:norms_and_margins}, Section~\ref{sub:exp_robust}).
According to the bound~\eqref{eq:robust_margin_bound}, one can assess expected adversarial error with $\epsilon$-bounded perturbations
by looking at the part of the plot to the right of~$\bar{\gamma} = 2\epsilon$.
In particular, the value of the CDF at such a value of~$\bar{\gamma}$ is representative of the
bound for large~$n$ (since the second term is negligible),
while for smaller~$n$, the best bound is obtained for a larger value of~$\bar{\gamma}$, which also suggests that
the right side of the plots is indicative of performance on small datasets.

When the RKHS norm can be well approximated, our bound provides a certificate on
test error in the presence of adversaries. 
While such an approximation is difficult to
obtain in general, the guarantee is most useful when lower and upper bounds of the RKHS norm are controlled together.

\subsection{New insights on generative adversarial networks}
\label{sub:gan_reg}

Generative adversarial networks (GANs) attempt to learn a \emph{generator} neural network~$G_\phi : \mathcal Z \to \mathcal X$,
so that the distribution of~$G_\phi(z)$ with~$z \sim D_z$ a noise vector resembles a data distribution~$D_x$.
In this section, we discuss connections between recent regularization techniques for
training GANs, and approaches to learning generative models
based on a MMD criterion~\citep{gretton2012kernel}, in view of our RKHS framework.
Our goal is to provide a new insight on these methods, but not necessarily to provide a new one.

Various recent approaches have relied on regularization strategies on a \emph{discriminator} network
in order to improve the stability of GAN training and the quality of the produced samples.
Some of these resemble the approaches presented in Section~\ref{sec:kernel_reg}
such as gradient penalties~\citep{gulrajani2017improved,roth2017stabilizing}
and spectral norm regularization~\citep[]{miyato2018spectral}.
We provide an RKHS interpretation of these methods as
optimizing an MMD distance with the convolutional kernel introduced in Section~\ref{sec:kernel_reg}:
\begin{equation}
\label{eq:ckn_mmd}
\min_\phi \sup_{\|f\|_\Hc \leq 1} \E_{x \sim D_x}[ f(x)] - \E_{z \sim D_z}[f(G_\phi(z))].
\end{equation}
When learning from an empirical distribution over~$n$ samples,
the MMD criterion is known to have much better sample complexity than the Wasserstein-1
distance considered by~\citet{arjovsky2017wasserstein} for high-dimensional data
such as images~\citep{sriperumbudur2012empirical}.
While the MMD approach has been used for training generative models, it generally relies on a generic kernel function,
such as a Gaussian kernel, that appears explicitly in the objective~\citep{dziugaite2015training,li2017mmd,binkowski2018demystifying}.
Although using a learned feature extractor can improve this, the Gaussian kernel might be a poor choice when
dealing with natural signals such as images, while the hierarchical kernel we consider in our paper is better suited
for this type of data, by providing useful invariance and stability properties.
Leveraging the variational form of the MMD~\eqref{eq:ckn_mmd} with this kernel suggests for instance using convolutional networks
as the discriminator~$f$, with constraints on the spectral norms in order to ensure~$\|f\|_\Hc \leq C$ for some~$C$,
as done by~\citet{miyato2018spectral} through normalization.

\section{Experiments}
\label{sec:experiments}

We tested the regularization strategies presented in Section~\ref{sec:kernel_reg}
in the context of improving generalization on small datasets and training robust models.
Our goal is to use common architectures used for large datasets and improve
their performance in different settings through regularization.
Our Pytorch implementation of the various strategies is available at \url{https://github.com/albietz/kernel_reg}.

For the adversarial training strategies, the inner maximization problems are solved using 5 steps of projected
gradient ascent with constant step-lengths.
In the case of the lower bound penalties $\|f\|_\delta^2$ and~$\|\nabla f\|^2$,
we also maximize over
examples in the mini-batch, only considering the maximal element when computing gradients with respect to parameters.
For the robust optimization problem~\eqref{eq:robust}, we use PGD with~$\ell_2$
perturbations,
as well as the corresponding~$\ell_2$ (squared) gradient norm penalty on the loss.
For the upper bound approaches with spectral norms (SNs), we consider the SN projection strategy with decaying~$\tau$,
as well as the SN penalty~\eqref{eq:optimization_problem_penalized}, either using power iteration (PI) or a full SVD
for computing gradients.

\subsection{Improving generalization on small datasets}
\label{sub:exp_smalldata}
We consider the datasets CIFAR10 and MNIST when using a small number of training examples,
as well as 102 datasets of biological sequences that suffer from small sample size.

\vspace*{-0.2cm}
\paragraph{CIFAR10.}
In this setting, we use 1\,000 and 5\,000 examples of the CIFAR10 dataset, with or without data augmentation.
We consider a VGG network~\citep{simonyan2014very} with 11 layers,
as well as a residual network~\citep{he2016deep} with 18 layers, which achieve
91\% and 93\% test accuracy respectively when trained on the full training set with standard data augmentation (horizontal flips + random crops).
We do not use batch normalization layers in order to prevent any interaction with spectral norms.
Each strategy derived in Section \ref{sec:kernel_reg} is trained for 500 epochs
using SGD with momentum and batch size 128, halving the step-size every 40 epochs.
In order to study the potential effectiveness of each method, we assume that a reasonably large validation set is
available to select hyper-parameters; thus, we keep 10\,000 annotated examples for this purpose.
We also show results using a smaller validation set in Appendix~\ref{sub:cifar_appx}.

\begin{table}[tb]
\caption{Regularization on CIFAR10 with 1\,000 examples for VGG-11 and ResNet-18.
Each entry shows the test accuracy with/without data augmentation when all hyper-parameters are optimized on a validation set.
See also Section~\ref{sub:cifar_appx} in the appendix for additional results and statistical testing.
}

\label{tab:smalldata}
\centering
\small
\vspace{0.2cm}
\begin{tabular}{ | l | c | c |  }
\hline
Method & 1k VGG-11 & 1k ResNet-18 \\ \hline
\hline
No weight decay & 50.70 / 43.75 & 45.23 / 37.12 \\
Weight decay & 51.32 / 43.95 & 44.85 / 37.09 \\
SN penalty (PI) & 54.64 / 45.06 & 47.01 / 39.63 \\
SN projection & 54.14 / \textbf{\color{darkgray}46.70} & 47.12 / 37.28 \\
VAT & 50.88 / 43.36 & 47.47 / 42.82 \\
PGD-$\ell_2$ & 51.25 / 44.40 & 45.80 / 41.87 \\
grad-$\ell_2$ & \textbf{\color{darkgray}55.19} / 43.88 & \textbf{49.30} / \textbf{\color{darkgray}44.65} \\
\hline
$\|f\|_\delta^2$ penalty & 51.41 / 45.07 & 48.73 / 43.72 \\
$\|\nabla f\|^2$ penalty & 54.80 / 46.37 & \textbf{\color{darkgray}48.99} / \textbf{44.97} \\
PGD-$\ell_2$ + SN proj & 54.19 / \textbf{\color{darkgray}46.66} & 47.47 / 41.25 \\
grad-$\ell_2$ + SN proj & \textbf{55.32} / \textbf{46.88} & 48.73 / 42.78 \\
$\|f\|_\delta^2$ + SN proj & 54.02 / \textbf{\color{darkgray}46.72} & 48.12 / 43.56 \\
$\|\nabla f\|^2$ + SN proj & \textbf{55.24} / \textbf{46.80} & \textbf{\color{darkgray}49.06} / \textbf{44.92} \\
\hline
\end{tabular}
\end{table}

Table~\ref{tab:smalldata} shows the test accuracies on 1\,000 examples for upper and lower bound approaches,
as well as combined ones.
We also include virtual adversarial training~\citep[VAT,][]{miyato2018virtual}.
We provide extended tables in Appendix~\ref{sub:cifar_appx} with additional methods,
other geometries, results for 5\,000 examples, as well as hypothesis tests for
comparing pairs of methods and assessing the significance of our findings.
Overall, we find that the combined lower bound + SN constraints approaches often
yield better results than either method separately.
For lower bound approaches alone, we found our $\|f\|_\delta^2$ and~$\|\nabla f\|^2$ penalties to often
work best, particularly without data augmentation,
while robust optimization strategies can be preferable with data augmentation,
perhaps thanks to the adaptive regularization effect discussed earlier, which may be helpful in
this easier setting.
Gradient penalties often outperform adversarial perturbation strategies,
possibly because of the closed form gradients which may improve optimization.
We also found that adversarial training strategies tend to poorly control SNs compared to
gradient penalties, particularly PGD (see also Section~\ref{sub:exp_robust}).
SN constraints alone can also work well in some cases, particularly for VGG architectures,
and often outperform SN penalties.
SN penalties can work well nevertheless and provide computational benefits when using the power iteration variant.

\begin{table}[tb]
\caption{Regularization on 300 or 1\,000 examples from MNIST, using deformations from Infinite MNIST.
($\ast$) indicates that random deformations were included as training examples,
while $\|f\|_\tau^2$ and $\|D_\tau f\|^2$
use them as part of the regularization penalty. 
See Section~\ref{sub:imnist_appx} in the appendix for more results and statistical testing.
}

\label{tab:imnist}
\centering
\small
\vspace{0.2cm}
\begin{tabular}{ | l | c | c |  }
\hline
Method & 300 VGG & 1k VGG \\ \hline
\hline
Weight decay & 89.32 & 94.08 \\
SN projection & 90.69 & 95.01 \\
grad-$\ell_2$ & 93.63 & 96.67 \\
$\|f\|_\delta^2$ penalty & 94.17 & 96.99 \\
$\|\nabla f\|^2$ penalty & 94.08 & 96.82 \\
\hline
Weight decay ($\ast$) & 92.41 & 95.64 \\
grad-$\ell_2$ ($\ast$) & 95.05 & 97.48 \\
\hline
$\|D_\tau f\|^2$ penalty & 94.18 & 96.98 \\
$\|f\|_\tau^2$ penalty & 94.42 & 97.13 \\
$\|f\|_{\tau}^2$ + $\|\nabla f\|^2$ & 94.75 & 97.40 \\
$\|f\|_{\tau}^2$ + $\|f\|^2_\delta$ & 95.23 & \textbf{\color{darkgray}97.66} \\
$\|f\|_{\tau}^2$ + $\|f\|^2_\delta$ ($\ast$) & \textbf{95.53} & \textbf{\color{darkgray}97.56} \\
$\|f\|_{\tau}^2$ + $\|f\|^2_\delta$ + SN proj & 95.20 & \textbf{\color{darkgray}97.60} \\
$\|f\|_{\tau}^2$ + $\|f\|^2_\delta$ + SN proj ($\ast$) & \textbf{\color{darkgray}95.40} & \textbf{97.77} \\
\hline
\end{tabular}
\end{table}

\vspace*{-0.2cm}
\paragraph{Infinite MNIST.}
In order to assess the effectiveness of lower bound penalties based on deformation stability,
we consider the Infinite MNIST dataset~\citep{loosli-canu-bottou-2006}, which provides an ``infinite''
number of transformed generated examples for each of the 60\,000 MNIST training digits.
Here, we use a 5-layer VGG-like network with average pooling after each 3x3 convolution layer,
in order to more closely match the architecture assumptions of~\citet{bietti2018group}
for deformation stability.
We consider two lower bound penalties that leverage the digit transformations in Infinite MNIST:
one based on ``adversarial'' deformations around each digit, denoted~$\|f\|_\tau^2$;
and a tangent propagation~\citep{simard1998transformation} variant, denoted $\|D_\tau f\|^2$,
which provides an approximation to~$\|f\|_\tau^2$ for small deformations
based on gradients along a few tangent vector directions given by deformations
(see Appendix~\ref{sec:deformation_penalties} for details).
Table~\ref{tab:imnist} shows the obtained test accuracy for subsets of MNIST of size 300 and 1\,000.
Overall, we find that combining both adversarial penalties $\|f\|_\tau^2$ and $\|f\|_\delta^2$
performs best, which suggests that it is helpful to obtain tighter lower approximations of the RKHS
norm by considering perturbations of different kinds.
Explicitly controlling the spectral norms can further improve performance,
as does training on deformed digits, which may yield better margins
by exploiting the additional knowledge that small deformations preserve labels.
Note that data augmentation alone (with some weight decay) does quite poorly in this case,
even compared to our lower bound penalties which do not use deformations.

\begin{table}[tb]
\caption{Regularization on protein homology detection tasks,
with or without data augmentation (DA).
Fixed hyperparameters are selected using the first half of the datasets,
and we report the average auROC50 score on the second half.
See Section~\ref{sub:protein_appx} in the appendix for more details and statistical testing.
}

\label{tab:protein}
\centering
\small
\vspace{0.2cm}
\begin{tabular}{|l|c|c|}
\hline
 Method                       &   No DA &    DA \\ \hline
\hline
 No weight decay              &   0.421 & 0.541 \\
 Weight decay                 &   0.432 & 0.544 \\
 SN proj                      &   0.583 & 0.615 \\
 PGD-$\ell_2$                 &   0.488 & 0.554 \\
 grad-$\ell_2$                &   0.551 & 0.570 \\ \hline
 $\|f\|_{\delta}^2$           &   0.577 & 0.611 \\
 $\|\nabla f\|^2$             &   0.566 & 0.598 \\
 PGD-$\ell_2$ + SN proj       &   \textbf{\color{darkgray}0.615} & \textbf{\color{darkgray}0.622} \\
 grad-$\ell_2$ + SN proj      &   0.581 & \textbf{\color{darkgray}0.634} \\
 $\|f\|_{\delta}^2$ + SN proj &   \textbf{0.631} & \textbf{0.639} \\
 $\|\nabla f\|^2$ + SN proj   &   0.576 & \textbf{\color{darkgray}0.617} \\
\hline
\end{tabular}
\vspace*{-0.1cm}
\end{table}

\vspace{-0.2cm}
\paragraph{Protein homology detection.}
Remote homology detection between protein sequences is an important problem to understand protein structure.
Given a protein sequence, the goal is to predict whether it belongs to a superfamily of interest.
We consider the Structural Classification Of Proteins (SCOP) version 1.67 dataset~\citep{murzin1995scop},
which we process as described in Appendix~\ref{sub:protein_appx}
in order to obtain 102 balanced binary classification tasks with 100 protein sequences each,
thus resulting in a low-sample regime. Protein sequences were also cut to $400$ amino acids.

Sequences are represented with a one-hot encoding strategy---that is, a
sequence of length~$l$ is represented as a binary matrix in $\{0,1\}^{20 \times
l}$, where $20$ is the number of different amino acids (alphabet size of the sequences).
Such a structure can then be processed by convolutional neural
networks~\citep{alipanahi2015predicting}.
In this paper, we do not try to optimize the structure of the network for the task,
since our goal is only to evaluate the effect of regularization strategies.
Therefore, we use a simple convolutional network with 3 convolutional layers
followed by global max-pooling and a final fully-connected layer
(we use filters of size 5, and a max-pooling layer after the second convolutional layer).

Training was done using Adam with a learning rate fixed to $0.01$, and a weight decay parameter
tuned for each method.
Since hyper-parameter selection per dataset is difficult due to the low sample size, we use the
same parameters across datasets. This allows us
to use the first 51 datasets as a validation set for hyper-parameter tuning,
and we report average performance with these fixed choices on the remaining 51 datasets.
The standard performance measure for this task is the auROC50 score (area under the ROC curve up to 50\% false positives).
We note that the selection of  hyper-parameters has a transductive component, since some of the sequences
in the test datasets may also appear in the datasets used for validation (possibly with a different label).

The results are shown in Table~\ref{tab:protein}.
The procedure used for data augmentation (right column) is described in Appendix~\ref{sub:protein_appx}.
We found that the most effective approach is the adversarial perturbation penalty,
together with SN constraints.
In particular, we found it to outperform the gradient penalty~$\|\nabla f\|^2$,
perhaps because in this case gradient penalties are only computed on a discrete set
of possible points given by one-hot encodings, while adversarial perturbations may
increase stability to wider regions, potentially covering different possible encoded
sequences.

\subsection{Training adversarially robust models}
\label{sub:exp_robust}

\begin{figure}[tb]
	\centering
	\includegraphics[width=.95\columnwidth]{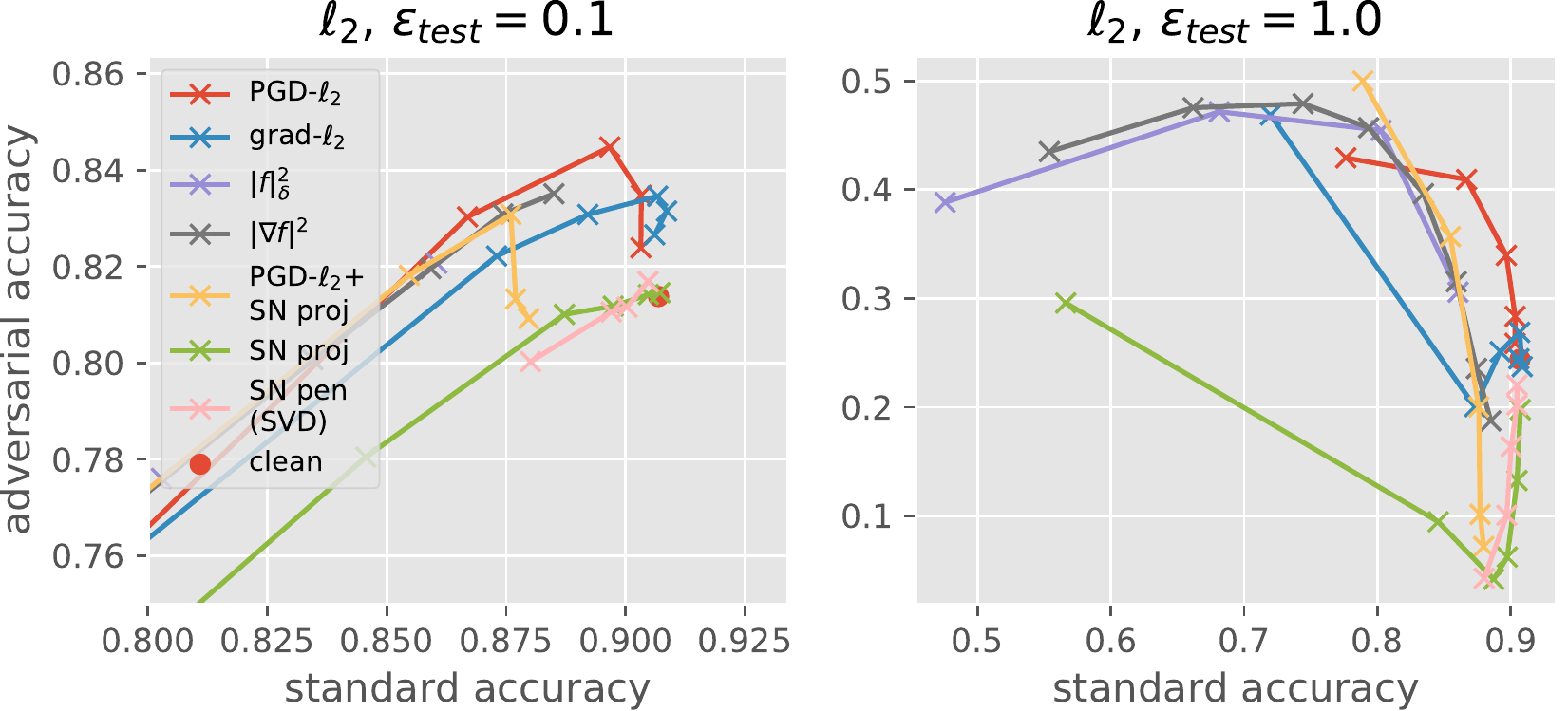}
        \vspace*{-0.1cm}
	\caption{Robustness trade-off curves of different regularization methods for VGG11 on CIFAR10.
	Each plot shows test accuracy vs adversarial test accuracy
	for $\ell_2$-bounded, 40-step PGD adversaries with a fixed~$\epsilon_{\text{test}}$.
	Different points on a curve correspond to training with different regularization strengths.
	The regularization increases monotonically along a given curve, and
	the leftmost points correspond to the strongest regularization.
	For PGD-$\ell_2$ + SN projection, we vary~$\epsilon$ with a fixed~$\tau = 0.8$.
	}
	\label{fig:robust_tradeoffs}
        \vspace*{-0.2cm}
\end{figure}

We consider the same VGG architecture as in Section~\ref{sub:exp_smalldata}, trained on CIFAR10
with data augmentation, with different regularization strategies.
Each method is trained for 300 epochs using SGD with momentum and batch size 128, dividing the step-size in half every 30 epochs.
This strategy was successful in reaching convergence for all methods.

Figure~\ref{fig:robust_tradeoffs} shows the test accuracy of the different methods in the presence
of $\ell_2$-bounded adversaries, plotted against standard accuracy.
We can see that the robust optimization approaches tend to work better in high-accuracy regimes, perhaps because the local stability that they encourage is sufficient on this dataset,
while the~$\|f\|_\delta^2$
penalty can be useful in large-perturbation regimes.
We find that upper bound approaches alone do not provide robust models,
but combining the SN constraint approach with a lower bound strategy
(in this case PGD-$\ell_2$) helps improve robustness perhaps thanks to a more explicit control of stability.
The plots also confirm that gradient penalties on the loss may be preferable for small regularization strengths (they achieve higher accuracy while improving robustness for small~$\epsilon_{test}$),
while for stronger regularization,
the gradient approximation no longer holds and the adversarial training approaches such
as PGD (and its combination with SN constraints) are preferred.
More experiments confirming these findings are available in Section~\ref{sub:robustness_appx} of the appendix.

\begin{figure}[tb]
	\centering
	\includegraphics[width=.48\columnwidth]{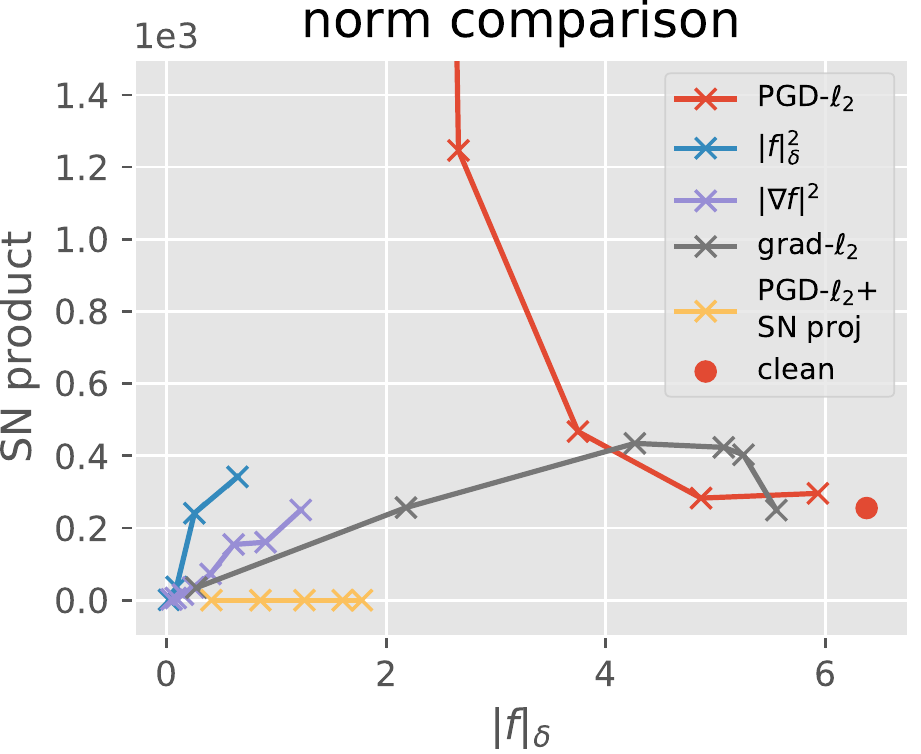}
	\includegraphics[width=.48\columnwidth]{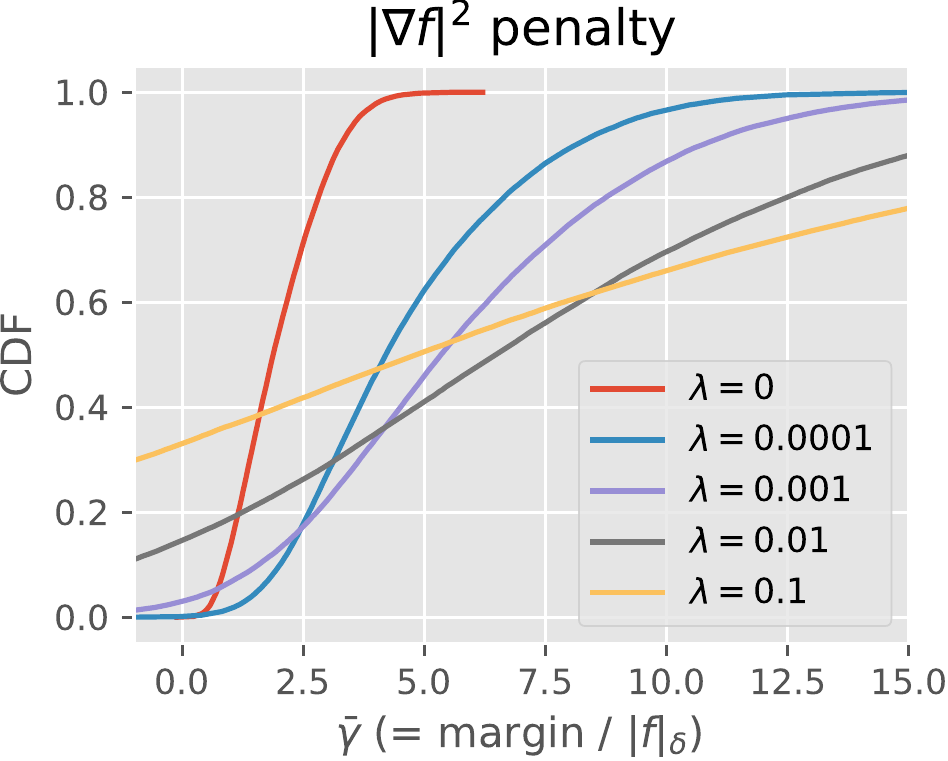}
        \vspace*{-0.1cm}
	\caption{(left) Comparison of lower and upper bound quantities ($\|f\|_\delta$ vs the product of spectral norms).
	(right) CDF plot of normalized empirical margins for the $\|\nabla f\|^2$ penalty with different
	regularization strengths, normalized by $\|f\|_\delta$.
	We consider 1000 fixed training examples when computing $\|f\|_\delta$.}
        \vspace*{-0.2cm}
	\label{fig:norms_and_margins}
\end{figure}

\vspace*{-0.2cm}
\paragraph{Norm comparison and adversarial generalization.}
Figure~\ref{fig:norms_and_margins} (left) compares lower and upper bound quantities
for different regularization strengths.
Note that for PGD, in contrast to other methods, we can see that the product of spectral norms (representative of an upper bound on~$\|f\|_\Hc$) increases when the lower bound~$\|f\|_\delta$ decreases.
This suggests that a network learned with PGD with large~$\epsilon$ may have large RKHS norm, possibly because the approach
tries to separate $\epsilon$-balls around the training examples,
which may require a more complex model than simply separating the training
examples~\citep[see also][]{madry2018towards}.
This large discrepancy between upper and lower bounds highlights the fact that such models may only
be stable locally near training data, though this happens to be enough for robustness on many test examples on CIFAR10.

In contrast, for other methods, and in particular the lower bound penalties~$\|f\|_\delta^2$ and~$\|\nabla f\|^2$,
the upper and lower bounds appear more tightly controlled, suggesting a more appropriate control
of the RKHS norm.
This makes our guarantees on adversarial generalization more meaningful,
and thus we may look at the empirical
distributions of normalized margins~$\bar{\gamma}$ obtained using $\|f\|_\delta$ for normalization (as an approximation of~$\|f\|_\Hc$),
shown in Figure~\ref{fig:norms_and_margins} (right).
The curves suggest that for small~$\bar{\gamma}$, and hence small $\epsilon_{test}$,
smaller values of~$\lambda$ are preferred, while stronger regularization helps
for larger~$\bar \gamma$, yielding lower test error guarantees in the presence of stronger adversaries
according to our bounds in
Section~\ref{sub:guarantees}.
This qualitative behavior is indeed observed in the results of Figure~\ref{fig:robust_tradeoffs} on test data
for the $\|\nabla f\|^2$ penalty.

\section*{Acknowledgements}
This work was supported by the ERC grant number 714381 (SOLARIS project) and by the MSR-Inria joint centre.

\bibliography{bibli}
\bibliographystyle{icml2019}

\newpage

\appendix
\onecolumn

Section~\ref{sec:experiments_appx} of this supplementary presents extended results
from our experiments, along with statistical tests for assessing the significance of our findings.
Section~\ref{sec:deformation_penalties} details our lower bound penalties based on deformations
and their relationship to tangent propagation.
Section~\ref{sec:spectral_norms_appx} presents our continuation algorithm for optimization
with spectral norm constraints.
Section~\ref{sec:non_euclidian_appx} describes
heuristic extensions of our lower bound regularization strategies to non-Euclidian geometries.
Finally, Section~\ref{sec:generalization_appx} provides our proof of the margin bound
of Proposition~\ref{prop:robust_margin_bound} for adversarial generalization.

\section{Additional Experiment Results}
\label{sec:experiments_appx}

\subsection{CIFAR10}
\label{sub:cifar_appx}

This section provides more extensive results for the experiments on CIFAR10 from Section~\ref{sub:exp_smalldata}.
In particular, Table~\ref{tab:smalldata_ext} shows additional experiments on larger subsets of size 5\.000,
as well as more methods, including different geometries (see Appendix~\ref{sec:non_euclidian_appx}).
The table also reports results obtained when using a smaller validation set of size 1\,000.
The full hyper-parameter grid is given in Table~\ref{tab:smalldata_param_grid}.

In order to assess the statistical significance of our results,
we repeated the experiments on 10 new random choices of subsets, using the hyperparameters selected
on the original subset from Table~\ref{tab:smalldata_ext} (except for learning rate, which is selected according to a different validation set for each subset).
We then compared pairs of methods using a paired t-test, with p-values shown in Table~\ref{tab:t_test}.
In particular, the results strengthen some of our findings, for instance, that $\|\nabla f \|^2$
should be preferred to the gradient penalty on the loss when there is no data augmentation,
and that combined upper+lower bound approaches tend to outperform the individual upper or lower
bound strategies.

\begin{table}[h]
\caption{Regularization on CIFAR10 with 1\,000 or 5\,000 examples for VGG-11 and ResNet-18.
Extended version of Table~\ref{tab:smalldata}.
Each entry shows the test accuracy with/without data augmentation when all hyper-parameters are optimized on a validation set of size 10\,000 (a) or 1\,000 (b),
and for the epoch with highest validation accuracy,
evaluating every 10 epochs (similar to early stopping).}
\centering
\vspace{0.2cm}
\label{tab:smalldata_ext}
(a) 10k examples in validation set

\begin{tabular}{ | l | c | c | c | c |  }
\hline
Method & 1k VGG-11 & 1k ResNet-18 & 5k VGG-11 & 5k ResNet-18 \\ \hline
\hline
No weight decay & 50.70 / 43.75 & 45.23 / 37.12 & 72.49 / 58.35 & 72.72 / 54.12 \\
Weight decay & 51.32 / 43.95 & 44.85 / 37.09 & 72.80 / 58.56 & 73.06 / 53.33 \\
SN penalty (PI) & 54.64 / 45.06 & 47.01 / 39.63 & 74.03 / 62.45 & 74.79 / 54.04 \\
SN penalty (SVD) & 53.44 / 46.06 & 47.26 / 37.94 & 74.53 / 62.93 & 75.59 / 54.98 \\
SN projection & 54.14 / \textbf{\color{darkgray}46.70} & 47.12 / 37.28 & 75.14 / 63.81 & 76.23 / 55.60 \\
VAT & 50.88 / 43.36 & 47.47 / 42.82 & 72.91 / 58.78 & 71.56 / 55.93 \\
PGD-$\ell_2$ & 51.25 / 44.40 & 45.80 / 41.87 & 73.18 / 58.98 & 72.53 / 55.92 \\
PGD-$\ell_\infty$ & 51.17 / 43.07 & 45.31 / 39.66 & 73.05 / 57.82 & 72.75 / 55.14 \\
grad-$\ell_2$ & \textbf{\color{darkgray}55.19} / 43.88 & \textbf{49.30} / \textbf{\color{darkgray}44.65} & \textbf{75.38} / 59.20 & 75.22 / 55.36 \\
grad-$\ell_1$ & 54.88 / 44.74 & \textbf{\color{darkgray}49.06} / 42.63 & \textbf{\color{darkgray}75.25} / 59.39 & 74.48 / 56.19 \\
\hline
$\|f\|_\delta^2$ penalty & 51.41 / 45.07 & 48.73 / 43.72 & 72.98 / 61.45 & 72.78 / 56.50 \\
$\|\nabla f\|^2$ penalty & 54.80 / 46.37 & \textbf{\color{darkgray}48.99} / \textbf{44.97} & 73.90 / 60.17 & 73.83 / \textbf{\color{darkgray}57.92} \\
PGD-$\ell_2$ + SN proj & 54.19 / \textbf{\color{darkgray}46.66} & 47.47 / 41.25 & 74.61 / \textbf{64.50} & \textbf{\color{darkgray}77.19} / 57.43 \\
grad-$\ell_2$ + SN proj & \textbf{55.32} / \textbf{46.88} & 48.73 / 42.78 & 75.11 / 63.54 & \textbf{77.73} / 57.09 \\
$\|f\|_\delta^2$ + SN proj & 54.02 / \textbf{\color{darkgray}46.72} & 48.12 / 43.56 & 74.55 / \textbf{\color{darkgray}64.33} & 75.64 / \textbf{59.03} \\
$\|\nabla f\|^2$ + SN proj & \textbf{55.24} / \textbf{46.80} & \textbf{\color{darkgray}49.06} / \textbf{44.92} & 72.31 / 63.74 & 72.24 / 57.56 \\
\hline
\end{tabular}
\\
\vspace{0.2cm}
(b) 1k examples in validation set

\begin{tabular}{ | l | c | c | c | c |  }
\hline
Method & 1k VGG-11 & 1k ResNet-18 & 5k VGG-11 & 5k ResNet-18 \\ \hline
\hline
No weight decay & 51.32 / 43.42 & 45.00 / 37.00 & 72.64 / 57.88 & 72.71 / 53.80 \\
Weight decay & 51.04 / 43.42 & 44.66 / 36.77 & 72.68 / 57.59 & 72.25 / 54.16 \\
SN penalty (PI) & 54.60 / 44.20 & 46.39 / 38.86 & 72.99 / 62.49 & 74.72 / 53.65 \\
SN penalty (SVD) & 53.76 / 44.79 & 47.31 / 37.92 & 74.05 / 63.34 & 75.73 / 54.65 \\
SN projection & 52.86 / \textbf{\color{darkgray}46.49} & 47.05 / 37.28 & 74.18 / \textbf{\color{darkgray}63.70} & 75.91 / 54.43 \\
VAT & 50.90 / 43.99 & 47.35 / 42.91 & 72.95 / 57.64 & 71.91 / 55.22 \\
PGD-$\ell_2$ & 50.95 / 43.26 & 45.77 / 41.71 & 72.71 / 57.68 & 72.87 / 54.17 \\
PGD-$\ell_\infty$ & 51.16 / 43.16 & 45.67 / 39.77 & 73.64 / 58.02 & 72.99 / 53.95 \\
grad-$\ell_2$ & \textbf{55.40} / 43.57 & 47.86 / \textbf{\color{darkgray}44.65} & \textbf{75.44} / 58.33 & 74.83 / 55.43 \\
grad-$\ell_1$ & 54.53 / 43.04 & \textbf{\color{darkgray}48.75} / 42.21 & \textbf{\color{darkgray}75.28} / 58.19 & 74.28 / 54.02 \\
\hline
$\|f\|_M^2$ penalty & 51.00 / 44.67 & 48.57 / 44.30 & 72.76 / 60.55 & 72.75 / 56.49 \\
$\|\nabla f\|^2$ penalty & 54.68 / 46.10 & 48.53 / \textbf{45.21} & 73.83 / 60.36 & 73.30 / \textbf{\color{darkgray}57.46} \\
PGD-$\ell_2$ + SN proj & 53.85 / \textbf{46.79} & 46.48 / 40.95 & 74.79 / 63.37 & \textbf{\color{darkgray}76.28} / \textbf{\color{darkgray}57.43} \\
grad-$\ell_2$ + SN proj & \textbf{\color{darkgray}55.28} / 45.11 & 48.42 / 41.93 & 75.17 / 63.45 & \textbf{77.24} / 56.18 \\
$\|f\|_M^2$ + SN proj & 54.00 / 45.14 & 47.12 / 41.86 & 74.54 / \textbf{63.94} & 75.25 / \textbf{57.94} \\
$\|\nabla f\|^2$ + SN proj & \textbf{\color{darkgray}55.21} / 45.68 & \textbf{49.03} / 43.58 & 71.92 / 63.47 & 71.83 / 56.06 \\
\hline
\end{tabular}
\end{table}

\begin{table}[h]
\caption{Paired t-tests comparing pairs of methods, on 10 different random choices of subsets of CIFAR10.
Each cell shows the p-value of the corresponding test, both with (left) and without (right) data augmentation.
We only show p-values smaller than~$0.05$.
Hyperparameters are fixed to the ones obtained for the results in Table~\ref{tab:smalldata}
(selected on a different choice of subset), except for the learning rate which is tuned on
a separate validation set for each choice of subset.}
\centering
\label{tab:t_test}
\vspace{0.2cm}
\begin{tabular}{ | c | c c | c c | c c | c c |  }
\hline
Test & \multicolumn{2}{|c|}{1k VGG-11} & \multicolumn{2}{|c|}{1k ResNet-18} & \multicolumn{2}{|c|}{5k VGG-11} & \multicolumn{2}{|c|}{5k ResNet-18} \\ \hline
\hline
SN projection $\succ$ Weight decay & 1e-04 & 1e-03 & - & - & 3e-06 & 1e-08 & 9e-07 & 4e-04\\ \hline
grad-$\ell_2$ $\succ$ Weight decay & 4e-09 & - & 2e-04 & 5e-05 & 7e-08 & 1e-04 & 5e-06 & -\\ \hline
$\|\nabla f\|^2$ $\succ$ Weight decay & 1e-08 & 2e-07 & 1e-05 & 3e-07 & 3e-04 & 5e-07 & 7e-03 & 1e-06\\ \hline
$\|\nabla f\|^2$ $\succ$ grad-$\ell_2$ & - & 3e-08 & 2e-02 & 2e-06 & - & 6e-05 & - & 4e-05\\ \hline
grad-$\ell_2$ $\succ$ $\|\nabla f\|^2$ & 2e-02 & - & - & - & 2e-05 & - & 7e-04 & -\\ \hline
grad-$\ell_2$ + SN proj $\succ$ grad-$\ell_2$ & - & 9e-03 & - & - & - & 5e-07 & 9e-06 & 2e-04\\ \hline
$\|\nabla f\|^2$ + SN proj $\succ$ $\|\nabla f\|^2$ & - & - & - & 1e-02 & - & 2e-06 & - & -\\ \hline
\end{tabular}
\end{table}

\begin{table}[h]
\caption{List of hyper-parameters used for each method on CIFAR10.
For each method, we additionally consider a learning rate parameter in $[0.003 ; 0.01 ; 0.03 ; 0.1]$.
For combined penalties, the sets of hyperparameters are listed in the same order as in the first column
(\ie, the choices of constraint radius are given last).}
\centering
\small
\vspace{0.2cm}
\label{tab:smalldata_param_grid}
\begin{tabular}{ | l | c |  }
\hline
Method & Parameter grid \\ \hline
\hline
No weight decay &  -   \\ \hline
Weight decay &  $[0; 0.0001 ; 0.0002 ; 0.0004 ; 0.0008 ; 0.001 ; 0.002]$   \\ \hline
SN penalty (PI) &  $[0.001 ; 0.003 ; 0.01 ; 0.03 ; 0.1 ; 0.3]$ \\ \hline
SN penalty (SVD) &  $[0.001 ; 0.003 ; 0.01 ; 0.03 ; 0.1 ; 0.3]$  \\ \hline
SN projection &  $[0.5 ; 0.6 ; 0.8 ; 1.0 ; 1.2 ; 1.4]$  \\ \hline
$\|f\|_\delta^2$ penalty & $[0.001 ; 0.003 ; 0.01 ; 0.03 ; 0.1]$  \\ \hline
$\|\nabla f\|^2$ penalty & $[0.00003 ; 0.0001 ; 0.0003 ; 0.001 ; 0.003 ; 0.01 ; 0.03]$ \\ \hline
VAT & $[0.1 ; 0.3 ; 1.0 ; 3.0]$ \\ \hline
PGD-$\ell_2$ &  $[0.003 ; 0.01 ; 0.03 ; 0.1 ; 0.3 ; 1.0]$ \\ \hline
PGD-$\ell_\infty$ &  $[0.001 ; 0.003 ; 0.01 ; 0.03 ; 0.1 ; 0.3]$  \\ \hline
grad-$\ell_1$ &  $[0.0001 ; 0.0003 ; 0.001 ; 0.003 ; 0.01 ; 0.03]$  \\ \hline
grad-$\ell_2$ &  $[0.001 ; 0.003 ; 0.01 ; 0.03 ; 0.1 ; 0.3 ; 1.0 ; 3.0]$ \\ \hline
PGD-$\ell_2$ + SN projection &  $ [0.003 ; 0.01 ; 0.03 ; 0.1] \times [0.6 ; 1.0 ; 1.4]$   \\ \hline
grad-$\ell_2$ + SN projection & $ [0.003 ; 0.01 ; 0.03 ; 0.1] \times [0.6 ; 1.0 ; 1.4]$  \\ \hline
$\|f\|_\delta^2$ + SN projection & $ [0.003 ; 0.01 ; 0.03] \times [0.6 ; 1.0 ; 1.4]$    \\ \hline
$\|\nabla f\|^2$ + SN projection & $ [0.001 ; 0.01 ; 0.1] \times [0.6 ; 1.0 ; 1.4]$   \\ \hline
\end{tabular}
\end{table}

\subsection{Infinite MNIST}
\label{sub:imnist_appx}

We provide more extensive results for the Infinite MNIST dataset in Table~\ref{tab:imnist_ext},
in particular showing more regularization strategies, as well as results with or without
data augmentation, marked with~$(\ast)$.
As in the case of CIFAR10, we use SGD with momentum (fixed to 0.9) for 500 epochs,
with initial learning rates in~$[0.005; 0.05; 0.5]$, and divide the step-size by 2 every 40 epochs.
The full hyper-parameter grid is given in Table~\ref{tab:imnist_param_grid}.

As in the case of CIFAR10, we report statistical significance tests in Table~\ref{tab:imnist_ttests} comparing pairs of
methods based on 10 different random choices of subsets.
In particular, the results confirm that weight decay with data augmentation alone
tends to give weaker results than separate penalties,
and that the combined penalty $\|f\|_\tau^2 + \|f\|_\delta^2$, which combines adversarial
perturbations of two different types,
outperforms each penalty taken by itself on a single type of perturbation,
which emphasizes the benefit of considering perturbations of different natures,
perhaps thanks to a tighter lower bound approximation of the RKHS norm.
We note that grad-$\ell_2 (\ast)$ worked well on some subsets,
but poorly on others due to training instabilities,
possibly because of the selected hyperparameters which are quite large
(and thus likely violate the approximation to the robust optimization objective).

\begin{table}
\caption{Test accuracies on subsets of MNIST using deformations from Infinite MNIST.
Extended version of Table~\ref{tab:imnist}.
($\ast$) indicates that random deformations were included as training examples (\ie, data augmentation),
while $\|f\|_\tau^2$ and $\|D_\tau f\|^2$
use them as part of the regularization penalty.
As in Table~\ref{tab:smalldata_ext}, we show results obtained using a validation set
of size 10\,000 (a) and 1\,000 (b).
}
\centering
\vspace{0.2cm}
\label{tab:imnist_ext}
\begin{tabular}{c c}
(a) 10k examples in validation set
& (b) 1k examples in validation set \\

\begin{tabular}{ | l | c | c |  }
\hline
Method & 300 VGG & 1k VGG \\ \hline
\hline
Weight decay & 89.32 & 94.08 \\
Weight decay ($\ast$) & 92.41 & 95.64 \\
SN projection & 90.69 & 95.01 \\
SN projection ($\ast$) & 92.17 & 95.88 \\
grad-$\ell_2$ & 93.63 & 96.67 \\
grad-$\ell_2$ ($\ast$) & 95.05 & 97.48 \\
\hline
$\|f\|_\delta^2$ penalty & 94.17 & 96.99 \\
$\|f\|_\delta^2$ penalty ($\ast$) & 94.86 & 97.40 \\
$\|\nabla f\|^2$ penalty & 94.08 & 96.82 \\
$\|\nabla f\|^2$ penalty ($\ast$) & 94.80 & 97.29 \\
$\|D_\tau f\|^2$ penalty & 94.18 & 96.98 \\
$\|D_\tau f\|^2$ penalty ($\ast$) & 94.91 & 97.29 \\
$\|f\|_\tau^2$ penalty & 94.42 & 97.13 \\
$\|f\|_\tau^2$ penalty ($\ast$) & 94.83 & 97.25 \\
$\|f\|_{\tau}^2$ + $\|\nabla f\|^2$ & 94.75 & 97.40 \\
$\|f\|_{\tau}^2$ + $\|\nabla f\|^2$ ($\ast$) & 95.14 & 97.44 \\
$\|f\|_{\tau}^2$ + $\|f\|^2_\delta$ & 95.23 & \textbf{\color{darkgray}97.66} \\
$\|f\|_{\tau}^2$ + $\|f\|^2_\delta$ ($\ast$) & \textbf{95.53} & \textbf{\color{darkgray}97.56} \\
grad-$\ell_2$ + SN proj & 93.89 & 96.85 \\
grad-$\ell_2$ + SN proj ($\ast$) & 95.15 & \textbf{97.80} \\
$\|f\|_\delta^2$ + SN proj & 93.97 & 96.89 \\
$\|f\|_\delta^2$ + SN proj ($\ast$) & 94.78 & 97.38 \\
$\|f\|_{\tau}^2$ + $\|\nabla f\|^2$ + SN proj & 95.09 & 97.42 \\
$\|f\|_{\tau}^2$ + $\|\nabla f\|^2$ + SN proj ($\ast$) & 95.03 & 97.27 \\
$\|f\|_{\tau}^2$ + $\|f\|^2_\delta$ + SN proj & 95.20 & \textbf{\color{darkgray}97.60} \\
$\|f\|_{\tau}^2$ + $\|f\|^2_\delta$ + SN proj ($\ast$) & \textbf{\color{darkgray}95.40} & \textbf{97.77} \\
\hline
\end{tabular}
&
\begin{tabular}{ | l | c | c |  }
\hline
Method & 300 VGG & 1k VGG \\ \hline
\hline
Weight decay & 89.32 & 93.34 \\
Weight decay ($\ast$) & 91.91 & 95.73 \\
SN projection & 90.60 & 94.83 \\
SN projection ($\ast$) & 92.01 & 95.91 \\
grad-$\ell_2$ & 92.92 & 96.42 \\
grad-$\ell_2$ ($\ast$) & \textbf{\color{darkgray}94.69} & \textbf{\color{darkgray}97.48} \\
\hline
$\|f\|_M^2$ penalty & 93.44 & 96.98 \\
$\|f\|_M^2$ penalty ($\ast$) & 94.57 & 97.14 \\
$\|\nabla f\|^2$ penalty & 94.08 & 96.77 \\
$\|\nabla f\|^2$ penalty ($\ast$) & 94.50 & 97.15 \\
$\|D_\tau f\|^2$ penalty & 94.03 & 97.16 \\
$\|D_\tau f\|^2$ penalty ($\ast$) & 94.15 & 96.64 \\
$\|f\|_\tau^2$ penalty & 93.53 & 97.13 \\
$\|f\|_\tau^2$ penalty ($\ast$) & \textbf{\color{darkgray}94.79} & 97.26 \\
$\|f\|_{\tau}^2$ + $\|\nabla f\|^2$ & \textbf{\color{darkgray}94.75} & 97.21 \\
$\|f\|_{\tau}^2$ + $\|\nabla f\|^2$ ($\ast$) & 94.43 & \textbf{\color{darkgray}97.42} \\
$\|f\|_{\tau}^2$ + $\|f\|^2_M$ & \textbf{95.15} & 97.27 \\
$\|f\|_{\tau}^2$ + $\|f\|^2_M$ ($\ast$) & \textbf{95.20} & \textbf{\color{darkgray}97.49} \\
grad-$\ell_2$ + SN proj & 93.44 & 96.81 \\
grad-$\ell_2$ + SN proj ($\ast$) & 94.05 & \textbf{97.60} \\
$\|f\|_M^2$ + SN proj & 93.97 & 96.61 \\
$\|f\|_M^2$ + SN proj ($\ast$) & \textbf{\color{darkgray}94.69} & 97.33 \\
$\|f\|_{\tau}^2$ + $\|\nabla f\|^2$ + SN proj & \textbf{\color{darkgray}94.75} & 97.16 \\
$\|f\|_{\tau}^2$ + $\|\nabla f\|^2$ + SN proj ($\ast$) & \textbf{\color{darkgray}94.74} & 97.22 \\
$\|f\|_{\tau}^2$ + $\|f\|^2_M$ + SN proj & \textbf{\color{darkgray}94.78} & \textbf{\color{darkgray}97.49} \\
$\|f\|_{\tau}^2$ + $\|f\|^2_M$ + SN proj ($\ast$) & \textbf{95.17} & \textbf{97.64} \\
\hline
\end{tabular}
\end{tabular}
\end{table}

\begin{table}
\caption{Paired t-tests comparing pairs of methods,
on 10 different random choices of subsets of MNIST.
Each cell shows the p-value of the corresponding test.
We only show p-values smaller than~$0.05$.
Hyperparameters are fixed to the ones obtained for the results in Table~\ref{tab:imnist}
(selected on a different choice of subset), except for the learning rate which is tuned on
a separate validation set for each choice of subset.
}
\centering
\vspace{0.2cm}
\label{tab:imnist_ttests}
\begin{tabular}{ | c | c | c |  }
\hline
Test & 300 VGG & 1k VGG \\ \hline
\hline
grad-$\ell_2$ ($\ast$) $\succ$ Weight decay ($\ast$) & - & 3e-11\\ \hline
$\|f\|_\tau^2$ penalty $\succ$ Weight decay ($\ast$) & 2e-08 & 2e-10\\ \hline
$\|f\|_{\tau}^2$ + $\|f\|^2_\delta$ $\succ$ Weight decay ($\ast$) & 1e-08 & 2e-10\\ \hline
$\|f\|_{\tau}^2$ + $\|f\|^2_\delta$ + SN proj ($\ast$) $\succ$ grad-$\ell_2$ ($\ast$) & - & 1e-02\\ \hline
grad-$\ell_2$ ($\ast$) $\succ$ $\|f\|_{\tau}^2$ + $\|f\|^2_\delta$ + SN proj ($\ast$) & - & -\\ \hline
$\|f\|_{\tau}^2$ + $\|f\|^2_\delta$ $\succ$ $\|f\|_\delta^2$ penalty & 1e-07 & 6e-09\\ \hline
$\|f\|_{\tau}^2$ + $\|f\|^2_\delta$ $\succ$ $\|f\|_\tau^2$ penalty & 2e-06 & 6e-07\\ \hline
$\|f\|_{\tau}^2$ + $\|f\|^2_\delta$ ($\ast$) $\succ$ $\|f\|_{\tau}^2$ + $\|f\|^2_\delta$ & 2e-03 & -\\ \hline
$\|f\|_{\tau}^2$ + $\|f\|^2_\delta$ + SN proj ($\ast$) $\succ$ $\|f\|_{\tau}^2$ + $\|f\|^2_\delta$ & 2e-03 & 2e-04\\ \hline
\end{tabular}
\end{table}

\begin{table}
\caption{List of hyper-parameters used for each method on Infinite MNIST.
For each method, we additionally consider a learning rate parameter in $[0.005 ; 0.05 ; 0.5]$.
For combined penalties, the sets of hyperparameters are listed in the same order as in the first column
(\eg, the choices of constraint radius are given last).}
\centering
\small
\vspace{0.2cm}
\label{tab:imnist_param_grid}
\begin{tabular}{ | l | c | }
\hline
Method & Grid \\ \hline
\hline
Weight decay & [0; 0.00001; 0.00003; 0.0001; 0.0003; 0.001; 0.003; 0.01; 0.03; 0.1] \\
SN projection & [1.0; 1.2; 1.4; 1.6; 1.8] \\
grad-$\ell_2$ & [0.1; 0.3; 1.0; 3.0; 10.0] \\
$\|f\|_\delta^2$ penalty & [0.1; 0.3; 1.0; 3.0] \\
$\|\nabla f\|^2$ penalty & [0.0003; 0.001; 0.003; 0.01; 0.03; 0.1; 0.3] \\
$\|D_\tau f\|^2$ penalty & [0.003; 0.01; 0.03; 0.1; 0.3] \\
$\|f\|_\tau^2$ penalty & [0.03; 0.1; 0.3; 1.0; 3.0] \\
$\|f\|_{\tau}^2$ + $\|\nabla f\|^2$ & [0.03; 0.1; 0.3; 1.0] $\times$ [0.003; 0.01; 0.03; 0.1] \\
$\|f\|_{\tau}^2$ + $\|f\|^2_\delta$ & [0.1; 0.3; 1.0] $\times$ [0.03; 0.1] \\
grad-$\ell_2$ + SN proj & [0.3; 1.0; 3.0; 10.0; 30.0] $\times$ [1.2; 1.6; 2.0] \\
$\|f\|_\delta^2$ + SN proj & [0.03; 0.1] $\times$ [1.2; 1.6; 2.0] \\
$\|f\|_{\tau}^2$ + $\|\nabla f\|^2$ + SN proj & [0.03; 0.1; 0.3] $\times$ [0.01; 0.03; 0.1] $\times$ [1.2; 1.6; 2.0] \\
$\|f\|_{\tau}^2$ + $\|f\|^2_\delta$ + SN proj & [0.1; 0.3; 1.0] $\times$ [0.03; 0.1] $\times$ [1.2; 1.6; 2.0] \\
\hline
\end{tabular}
\end{table}

\subsection{Protein homology detection}
\label{sub:protein_appx}

\paragraph{Dataset description.}
Our protein homology detection experiments consider
the Structural Classification Of Proteins (SCOP) version 1.67 dataset \citep{murzin1995scop},
filtered and split following the procedures of \cite{haandstad2007motif}.
Specifically, positive training samples are extracted from one superfamily from which one family is withheld to serve as positive test set, while negative sequences are chosen from outside of the target family’s hold and are randomly split into training and test samples in the same ratio as positive samples.
This yields 102 superfamily classification tasks, which are generally very class-imbalanced.
For each task, we sample 100 class-balanced training samples to use as training set. The positive samples are extended to 50 with Uniref50 using PSI-BLAST \citep{altschul1997gapped} if they are fewer.

\paragraph{Data augmentation procedure.}
We consider in our experiments a discrete way of perturbing training samples to 
perform data augmentation. Specifically, for a given sequence, a perturbed sequence 
can be obtained by randomly changing some of the characters. Each character in the sequence 
is switched to a different one, randomly chosen from the alphabet, with some 
probability $p$. We fixed this probability to 0.1 throughout the experiments.

\paragraph{Experimental details and significance tests.}
In our experiments, we use the Adam optimization algorithm with a learning rate fixed
to 0.01 (and $\beta$ fixed to defaults $(0.9, 0.999)$),
with a batch size of 100 for 300 epochs.
The full hyper-parameter grid is given in Table~\ref{tab:protein_param_grid}.
In addition to the average auROC50 scores reported in Table~\ref{tab:protein},
we perform paired t-tests for comparing pairs of methods in Table~\ref{tab:protein_ttests}
in order to verify the significance of our findings.
The results confirm that the adversarial perturbation penalty and its combination
with spectral norm constraints tends to outperform the other approaches.

\begin{table}
\caption{Paired t-tests comparing pairs of methods on the 51 test datasets from the
set of protein homology detection tasks.
Each cell shows the p-value of the corresponding test.
We only show p-values smaller than~$0.05$.
We use the same hyperparameters as the ones obtained in the results of Table~\ref{tab:protein}.
}
\centering
\vspace{0.2cm}
\label{tab:protein_ttests}
\begin{tabular}{|l|c|c|}
\hline
 Test                                                            &   No DA &      DA \\ \hline
\hline
 SN proj $\succ$ Weight decay                                    &   1e-05 &   4e-05 \\
 grad-$\ell_2$ $\succ$ Weight decay                              &   5e-05 &   5e-02 \\
 $\|f\|_{\delta}^2$ $\succ$ Weight decay                         &   5e-06 &   3e-05 \\
 $\|\nabla f\|^2$ $\succ$ Weight decay                           &   9e-06 &   3e-03 \\
 $\|f\|_{\delta}^2$ $\succ$ grad-$\ell_2$                        &   -     &   4e-03 \\
 $\|\nabla f\|^2$ $\succ$ grad-$\ell_2$                          &   -     &   -     \\
 grad-$\ell_2$ + SN proj $\succ$ grad-$\ell_2$                   &   -     &   1e-03 \\
 $\|f\|_{\delta}^2$ + SN proj $\succ$ $\|f\|_{\delta}^2$         &   3e-03 &   5e-02 \\
 $\|\nabla f\|^2$ + SN proj $\succ$ $\|\nabla f\|^2$             &   -     &   -     \\
 $\|f\|_{\delta}^2$ + SN proj $\succ$ $\|\nabla f\|^2$ + SN proj &   8e-05 &   -     \\
\hline
\end{tabular}
\end{table}

\begin{table}
\caption{List of hyper-parameters used for each method on protein homology detection datasets.
For combined penalties, the hyperparameters are the cross-products of each individual method.}
\centering
\small
\vspace{0.2cm}
\label{tab:protein_param_grid}
\begin{tabular}{|c|c|}
\hline
 Method                                                       &   Parameter grid \\ \hline
\hline
 No weight decay  &  $-$                          \\
 Weight decay     &  $[0; 0.01; 0.001; 0.0001; 0.00001]$ \\
 SN proj          &  $[10; 1.0; 0.1]$             \\
 PGD-$\ell_2$     &  $[100.0; 10.0; 1.0; 0.1]$    \\
 grad-$\ell_2$    &  $[100.0; 10.0; 1.0; 0.1; 0.01, 0.001]$     \\
 $\|f\|_{\delta}^2$      &  $[10.0; 1.0; 0.1]$           \\
 $\|\nabla f\|^2$ &  $[10.0; 1.0; 0.1; 0.01; 0.001; 0.0001]$ \\
\hline
\end{tabular}
\end{table}

\subsection{Robustness}
\label{sub:robustness_appx}

\begin{figure*}
	\centering
	\includegraphics[width=.9\textwidth]{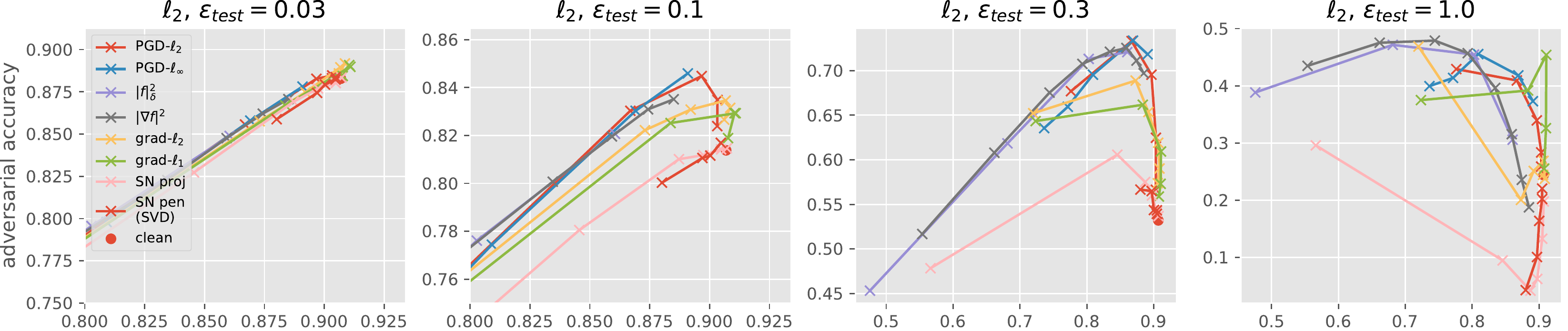}
	\includegraphics[width=.9\textwidth]{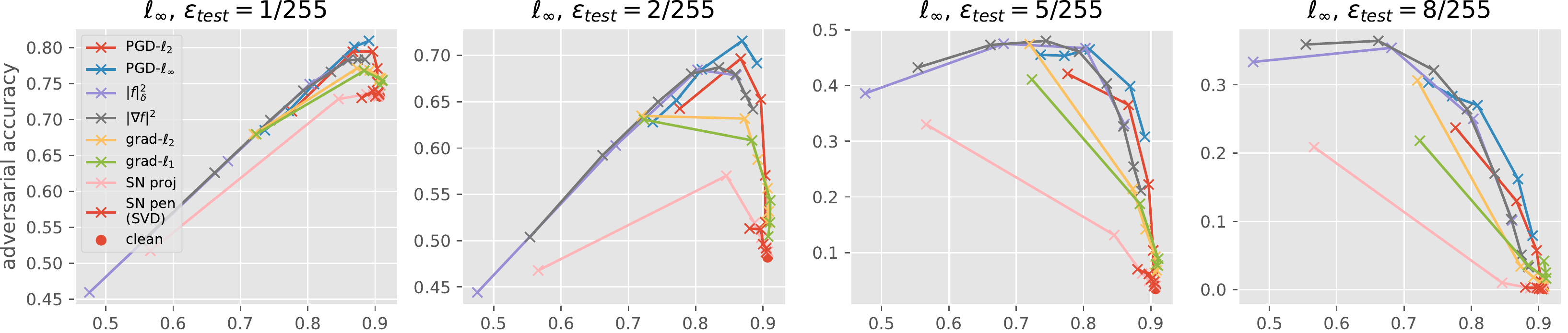}
	\includegraphics[width=.9\textwidth]{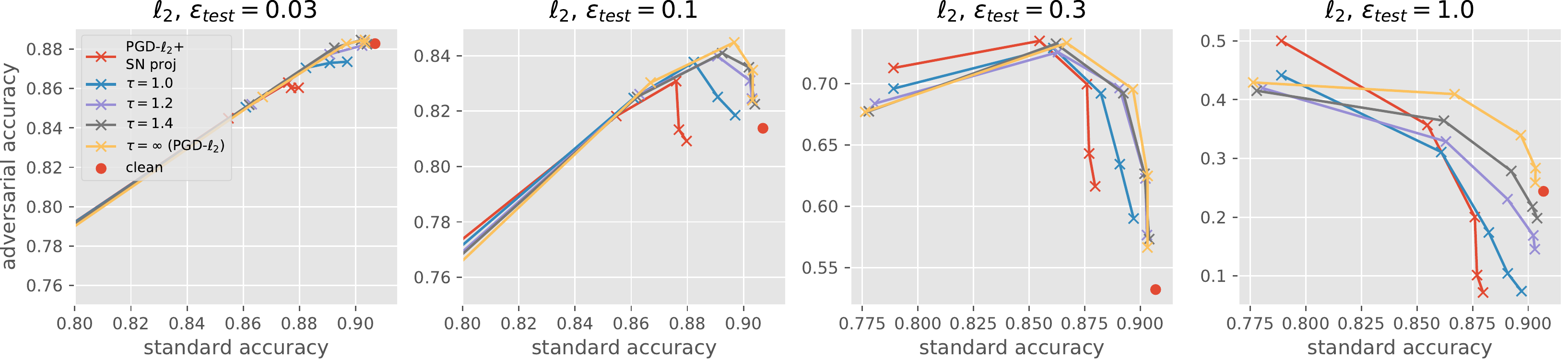}
	\caption{Robustness trade-off curves of different regularization methods for VGG11 on CIFAR10 (extended
	version of Figure~\ref{fig:robust_tradeoffs}).
	The plots show test accuracy vs adversarial test accuracy
	for $\ell_2$-bounded (top/bottom) or $\ell_\infty$-bounded (middle),
	40-step PGD adversaries with a fixed~$\epsilon_{\text{test}}$.
	Different points on a curve correspond to training with different regularization strengths.
	The regularization increases monotonically along a given curve, and
	the leftmost points correspond to the strongest regularization.
	The bottom plots consider PGD-$\ell_2$ + SN projection,
	with different fixed values of the constraint radius~$\tau$, for varying $\epsilon$ in PGD.}
	\label{fig:robust_tradeoffs_appx}
\end{figure*}

Figure~\ref{fig:robust_tradeoffs_appx} extends Figure~\ref{fig:robust_tradeoffs}
from Section~\ref{sub:exp_robust} to show more methods, adversary strenghts, and different geometries.
For combined (PGD-$\ell_2$ + SN projection) approaches, we can see that stronger constraints (\ie, smaller~$\tau$)
tend to reduce standard accuracy, likely because it prevents a good fit of the data,
but can provide better robustness to strong adversaries ($\epsilon_{test} = 1$).
We can see that using the right metric in PGD indeed helps against an~$\ell_\infty$ adversary,
nevertheless controlling global stability through the RKHS norm as in the~$\|f\|_\delta^2$ and~$\|\nabla f\|^2$
penalties can still provide some robustness against such adversaries, even with large $\epsilon_{test}$.
For gradient penalties, we find that the different geometries behave quite similarly,
which may suggest that more appropriate optimization algorithms than SGD could be needed to
better accommodate the non-smooth case of $\ell_1/\ell_\infty$, or perhaps that both algorithms are actually
controlling the same notion of complexity on this dataset.

\clearpage

\section{Details on Deformation Stability Penalties}
\label{sec:deformation_penalties}

This section provides more details on the deformation stability penalties mentioned in
Section~\ref{sub:lower_bounds}, and the practical versions we use in our experiments on the
Infinite MNIST dataset~\citep{loosli-canu-bottou-2006}.

\paragraph{Stability to deformations.}
We begin by providing some background on deformation stability,
recalling that these can provide new lower bound penalties as explained in Section~\ref{sub:lower_bounds}.
Viewing an element $x \in \mathcal X$ as a signal $x(u)$, where $u$ denotes the location
(\eg~a two-dimensional vector for images),
we denote by $x_\tau$ a deformed version of~$x$ given by $x_\tau(u) = x(u - \tau(u))$,
where~$\tau$ is a diffeomorphism.
The deformation stability bounds of~\citet{bietti2018group} take the form:
\begin{equation}
\label{eq:stability_bound}
\|\Phi(x_\tau) - \Phi(x)\|_\Hcal \leq (C_1 \|\tau\|_\infty + C_2 \|\nabla \tau\|_\infty) \|x\|,
\end{equation}
where $\nabla \tau (u)$ is the Jacobian of $\tau$ at location~$u$.
Here, $C_1$ controls translation invariance and typically decreases with the total amount of pooling
(\ie, translation invariance more or less corresponds to the resolution at the final layer),
while~$C_2$ controls stability to deformations (note that $\nabla \tau = 0$ for translations)
and is typically smaller when using small patches.
We note that the bounds assume linear pooling layers with a certain spatial decay,
adapted to the resolution of the current layer;
our experiments on Infinite MNIST with deformation stability penalties
thus use average pooling layers on 2x2 neighborhoods.

\paragraph{Adversarial deformation penalty.}
We can obtain lower bound penalties by exploiting the above stability bounds in
a similar manner to the adversarial perturbation penalty introduced in Section~\ref{sub:lower_bounds}.
In particular, assuming a scalar-valued convolutional network~$f$:
\begin{equation}
\label{eq:adv_deformation}
\|f\|_\tau^2 := \sup_{x \in \mathcal X, \tau \in \mathcal T} (f(x_\tau) - f(x))^2 \\
\end{equation}
where~$\mathcal T$ is a collection of diffeomorphisms.
When the diffeomorphisms in~$\mathcal T$ have bounded norm~$\|\tau\|_\infty$ and
Jacobian norm~$\|\nabla \tau\|_\infty$,
and assuming $\mathcal X$ (or, in practice, the training data) is bounded,
the stability bound~\ref{eq:stability_bound} ensures that
the set $U_{\mathcal T} = \{\Phi(x_\tau) - \Phi(x) : x \in \mathcal X, \tau \in \mathcal T\}$ is included in an
RKHS ball with some radius $r$, so that~$\|f\|_\tau$ is a lower bound on~$r \|f\|_\Hcal$.

\paragraph{Tangent gradient penalty.}
We also consider the following gradient penalty along tangent vectors,
which provides an approximation of the above adversarial penalty when
considering small, parameterized deformations,
and recovers the tangent propagation strategy of~\citet{simard1998transformation}:
\begin{equation}
\label{eq:tangent_gradient}
\|D_\tau f\|^2 := \sup_{x \in \mathcal X} \|\partial_\alpha f(x + \sum_i \alpha_i t_{x,i}) \|^2,
\end{equation}
where $\{t_{x,i}\}_{i=1,\ldots, q}$ are
tangent vectors at~$x$ obtained from a given set of deformations.
To see the link with the adversarial deformation penalty~\ref{eq:adv_deformation},
consider for simplicity a single deformation,~$\mathcal T = \{\tau_0\}$.
For small~$\alpha$, we have
\begin{align*}
x_{\alpha \tau_0} \approx x + \alpha t_x, \quad \text{where} \quad t_x(u) = \tau_0(u) \cdot \nabla x(u),
\end{align*}
where~$t_x$ denotes the tangent vector of the deformation
manifold $\{\alpha \tau_0 : \alpha\}$ at~$\alpha = 0$~\citep{simard1998transformation}.
Then,
\[
f(x_{\alpha\tau_0}) - f(x) \approx \alpha \partial_\alpha f(x + \alpha t_x) = \alpha \langle \nabla f(x), t_x \rangle.
\]
In this case, denoting $\alpha \mathcal T = \{\alpha \tau_0\}$, we have
\[
\sup_{x \in \mathcal X, \tau \in \alpha \mathcal T} (f(x_\tau) - f(x))^2 \approx \alpha^2 \sup_{x \in \mathcal X} |\partial_\alpha f(x + \alpha t_x)|^2,
\]
so that when $\alpha$ is small, the adversarial penalty can be approximated by $\alpha \|D_\tau f\|$
(note that using $\alpha \mathcal T$ instead of~$\mathcal T$ in the adversarial penalty
would also yield a scaling by~$\alpha$, since the stability bounds imply $\alpha$ times smaller
perturbations in the RKHS).

\paragraph{Practical implementations on Infinite MNIST.}
In our experiments on Infinite MNIST, we compute $\|f\|_\tau^2$ by considering 32 random transformations
of each digit in a mini-batch of training examples,
and taking the maximum over both the example and the transformation.
We do this separately for each class, as for the other lower bound penalties $\|f\|_\delta^2$ and $\|\nabla f\|^2$.
For~$\|D_\tau f\|^2$, we take $\{t_{x,i}\}_{i=1,\ldots,q}$ with~$q=30$ to be tangent vectors given
by random diffeomorphisms from Infinite MNIST around each example~$x$.

\section{Details on Optimization with Spectral Norms}
\label{sec:spectral_norms_appx}

\label{sub:projected_sgd}

This section details our optimization approach presented in Section~\ref{sub:upper_bounds}
for learning with spectral norm constraints.
In particular, we rely on a \emph{continuation} approach, decreasing the size of the ball constraints
during training, towards a final value~$\tau$. The method is presented in Algorithm~\ref{alg:psgd}.
We use an exponentially decreasing schedule for $\tau$,
and take $\kappa$ to be 2 epochs for regularization, and 50 epochs for robustness.
In the context of convolutional networks, we simply consider the SVD of a reshaped filter matrix,
but we note that alternative approaches based on the singular values of the full convolutional operation
may also be used~\citep{sedghi2018singular}.

\begin{algorithm}[th]
	\caption{Stochastic projected gradient with continuation}
	\label{alg:psgd}
	\begin{algorithmic}
	\STATE Input: $\tau$, $\kappa$, step-sizes $\eta_t$
		\FOR{$t = 1, \ldots$}
		\STATE Sample mini-batch and compute gradients of the loss w.r.t. each $W^l$, denoted~$G_t^l$
		\STATE $\tau_{t}=\tau (1 + \exp{\left(\frac{-t}{\kappa}\right)})$
		\FOR{$l = 1, \ldots, L$}
		\STATE 	$\tilde W_t^l := W_t^l - \eta_t G_t^l$
		\STATE Compute SVD: $\tilde W_t^l = U \text{diag}(\sigma) V^T$
		\STATE  $ \widehat{\sigma} := \text{proj}_{\|.\|_{\infty} \leq \tau_t}\left(\sigma\right)$
		\STATE 	$ W_{t+1}^l := U\text{diag}(\widehat{\sigma})V^T$
		\ENDFOR
		\ENDFOR
	\end{algorithmic}
\end{algorithm}


\section{Extensions to Non-Euclidian Geometries}
\label{sec:non_euclidian_appx}


The kernel approach from previous sections is well-suited for input spaces~$\mathcal X$ equipped with the Euclidian
distance, thanks to the non-expansiveness property~\eqref{eq:non_expansive} of the kernel mapping.
In the case of linear models, this kernel approach corresponds to using $\ell_2$-regularization by taking a linear kernel.
However, other forms of regularization and geometries can often be useful,
for example to encourage sparsity with an~$\ell_1$ regularizer.
Such a regularization approach presents tight links with robustness to~$\ell_\infty$ perturbations on input data,
thanks to the duality relation $\|w\|_1 = \sup_{\|u\|_\infty} \langle w, u \rangle$~\citep[see][]{xu2009robust}.

In the context of deep networks, we can leverage such insights to obtain new regularizers,
expressed in the same variational form as the lower bounds in Section~\ref{sub:lower_bounds},
but with different geometries on~$\mathcal X$. For $\ell_\infty$ perturbations, we obtain
\begin{equation}
\label{eq:gradient_l1}
\sup_{x, y\in \mathcal X} \frac{f(x) - f(y)}{\|x - y\|_\infty} \quad \geq \quad \sup_{x \in \mathcal X} \|\nabla f(x) \|_1.
\end{equation}
The Lipschitz regularizer (l.h.s.) can also be taken in an adversarial perturbation form, with~$\ell_\infty$-bounded perturbations $\|\delta\|_\infty \leq \epsilon$.
When considering the corresponding robust optimization problem
\begin{equation}
\label{eq:robust_linf}
\min_\theta \frac{1}{n} \sum_{i=1}^n \sup_{\|\delta\|_\infty \leq \epsilon} \ell(y_i, f_\theta(x_i + \delta)),
\end{equation}
we may consider the PGD approach of~\citet{madry2018towards}, or the associated gradient penalty
approach with the~$\ell_1$ norm, which is a good approximation when~$\epsilon$ is small~\citep{lyu2015unified,simon2018adversarial}.

As most visible in the gradient $\ell_1$-norm in~\eqref{eq:gradient_l1}, these penalties encourage some sparsity
in the gradients of~$f$, which is a reasonable prior for regularization on images, for instance, where we might only
want predictions to change based on few salient pixel regions. This can lead to gains in interpretability, as observed by~\citet{tsipras2018there}.

We note that in the case of linear models, our robust margin bound of Section~\ref{sub:guarantees} can be adapted to
$\ell_\infty$-perturbations, by leveraging Rademacher complexity bounds for $\ell_1$-constrained models~\citep{kakade2009complexity}.
Obtaining similar bounds for neural networks would be interesting but goes beyond the scope of this paper.



\section{Details on Generalization Guarantees}
\label{sec:generalization_appx}

This section presents the proof of Proposition~\ref{prop:robust_margin_bound},
which relies on standard tools from statistical learning theory~\citep[\eg,][]{boucheron2005theory}.

\subsection{Proof of Proposition~\ref{prop:robust_margin_bound}}
\begin{proof}
Assume for now that~$\gamma$ is fixed in advance, and let $\mathcal F_\lambda := \{f \in \Hc : \|f\|_\Hc \leq \lambda\}$.
Note that for all~$f \in \mathcal F_\lambda$ we have
\begin{align*}
\text{err}_\mathcal{D}(f, \epsilon) = P(\exists \|\delta\| \leq \epsilon: y f(x + \delta) < 0) \leq P(yf(x) < \lambda \epsilon) =: L^{\lambda \epsilon}(f),
\end{align*}
since~$\|f\|_\Hc \leq \lambda$ is an upper bound on the Lipschitz constant of~$f$.
Consider the function
\begin{align*}
\phi(x) = \begin{cases}
	0, &\text{ if }x \leq -\gamma - \lambda \epsilon\\
	1, &\text{ if }x \geq - \lambda \epsilon\\
	1 + (x + \lambda \epsilon)/\gamma, &\text{ otherwise.}
\end{cases}
\end{align*}
Defining $A(f) = \E \phi(-y f(x)) \geq L^{\lambda \epsilon}(f)$ and $A_n(f) = \frac{1}{n} \sum_{i=1}^n \phi(- y_i f(x_i)) \leq L_n^{\lambda \epsilon + \gamma}(f)$,
and noting that $\phi$ is upper bounded by 1 and $1/\gamma$ Lipschitz,
we can apply similar arguments to~\citep[Theorem 4.1]{boucheron2005theory} to obtain,
with probability $1 - \delta$,
\begin{equation*}
L^\lambda \epsilon(f) \leq L_n^{\lambda \epsilon + \gamma}(f) + O \left(\frac{1}{\gamma} R_n(\mathcal{F}_\lambda) + \sqrt{\frac{\log 1/\delta}{n}} \right),
\end{equation*}
where~$R_n(\mathcal{F}_\lambda)$ denotes the empirical Rademacher complexity of~$\mathcal{F}_\lambda$ on the dataset $\{(x_i, y_i)\}_{i=1, \ldots, n}$.
Standard upper bounds on empirical Rademacher complexity of kernel classes with bounded RKHS norm yield the following bound
\begin{align*}
\text{err}_\mathcal{D}(f, \epsilon) \leq L_n^{\lambda \epsilon + \gamma}(f) + O \left( \frac{\lambda}{\gamma \sqrt{n}} \sqrt{\frac{1}{n}\sum_{i=1}^n K(x_i, x_i)} + \sqrt{\frac{\log 1/\delta}{n}} \right).
\end{align*}
Note that the bound is still valid with $\gamma' \geq \gamma$ instead of~$\gamma$ in the first term
of the r.h.s., since $L_n^{\gamma}(f)$ is non-decreasing as a function of~$\gamma$.

In order to establish the final bound, we instantiate the previous bound for values $\lambda_i = 2^i$ and $\gamma_j = 2^{-j}$.
Defining $\delta_{i,j} = \frac{\delta}{(1 + 4i^2) \cdot ( 1 + 4j^2)}$, we have that w.p. $1 - \delta_{i,j}$, for all $f \in \mathcal F_{\lambda_i}$ and all $\gamma \geq \gamma_j$,

\begin{align}
\label{eq:bound_single}
\text{err}_\mathcal{D}(f, \epsilon) \leq L_n^{\lambda_i \epsilon + \gamma}(f) + O \left( \frac{\lambda_i}{\gamma_j \sqrt{n}} \sqrt{\frac{1}{n}\sum_{i=1}^n K(x_i, x_i)} + \sqrt{\frac{\log 1/\delta_{i,j}}{n}} \right).
\end{align}
By a union bound, this event holds jointly for all integers $i, j$ w.p. greater than $1 - \delta$,
since $\sum_{i,j} \delta_{i,j} \leq \delta$.
Now consider an arbitrary $f \in \Hc$ and $\gamma > 0$ and let $i = \lceil \log_2 \|f\|_\Hc \rceil$
and $j = \lceil \log_2 (1/\gamma) \rceil$. We have
\begin{align*}
\lambda_i &\leq 2 \|f\|_\Hc \\
\frac{1}{\gamma_j} &\leq \frac{2}{\gamma} \\
\log(1/\delta_{i,j}) &\leq \log(C(\|f\|_\Hc, \gamma) / \delta),
\end{align*}
with $C(\|f\|_\Hc, \gamma) := (1 + 4(\log_2\|f\|_\Hc)^2) \cdot (1 + 4(\log_2 (1/\gamma))^2)$.
Applying this to the bound in~\eqref{eq:bound_single} yields the desired result.

\end{proof}

\end{document}